\documentclass[10pt,twocolumn,letterpaper]{article}

\usepackage[pagenumbers]{cvpr} 
\usepackage[accsupp]{axessibility}  

\definecolor{cvprblue}{rgb}{0.21,0.49,0.74}
\usepackage[pagebackref,breaklinks,colorlinks,allcolors=cvprblue]{hyperref}
\usepackage{booktabs}
\usepackage{graphicx}
\usepackage[table,xcdraw]{xcolor}
\usepackage{multirow,multicol}
\usepackage{makecell}
\usepackage{titletoc}
\def\paperID{} 
\def\confName{CVPR}
\def\confYear{2026}

\definecolor{myblue}{RGB}{173,216,230} 

\def\darkred{\color{red!80!black}}

\title{ShowTable: Unlocking Creative Table Visualization with Collaborative Reflection and Refinement}

\author{
Zhihang Liu\textsuperscript{1}\thanks{Equal contribution, interns at Tongyi Lab.},
Xiaoyi Bao\textsuperscript{2}\footnotemark[1],
Pandeng Li\textsuperscript{1,7}\thanks{Corresponding author.},
Junjie Zhou\textsuperscript{3},
Zhaohe Liao\textsuperscript{4},
Yefei He\textsuperscript{5}, \\
Kaixun Jiang\textsuperscript{6},
Chen-Wei Xie\textsuperscript{7},
Yun Zheng\textsuperscript{7},
Hongtao Xie\textsuperscript{1}
\\[0.5ex]
\textsuperscript{1~}USTC\quad
\textsuperscript{2~}CASIA\quad
\textsuperscript{3~}NJU\quad
\textsuperscript{4~}SJTU\quad
\textsuperscript{5~}ZJU\quad
\textsuperscript{6~}FDU\quad
\textsuperscript{7~}Tongyi Lab\quad
\\[0.1ex]
{\tt\small Project Page: \url{https://lntzm.github.io/showtable-page/}}
}

\begin{document}
\maketitle

\begin{abstract}
While existing generation and unified models excel at general image generation, they struggle with tasks requiring deep reasoning, planning, and precise data-to-visual mapping abilities beyond general scenarios. To push beyond the existing limitations, we introduce a new and challenging task: creative table visualization, requiring the model to generate an infographic that faithfully and aesthetically visualizes the data from a given table. To address this challenge, we propose ShowTable, a pipeline that synergizes MLLMs with diffusion models via a progressive self-correcting process. The MLLM acts as the central orchestrator for reasoning the visual plan and judging visual errors to provide refined instructions, the diffusion execute the commands from MLLM, achieving high-fidelity results. To support this task and our pipeline, we introduce three automated data construction pipelines for training different modules. Furthermore, we introduce TableVisBench, a new benchmark with 800 challenging instances across 5 evaluation dimensions, to assess performance on this task. Experiments demonstrate that our pipeline, instantiated with different models, significantly outperforms baselines, highlighting its effective multi-modal reasoning, generation, and error correction capabilities.
\end{abstract}

\section{Introduction}
\label{sec:intro}

\begin{figure}[!t]
    \centering
    \includegraphics[width=0.49\textwidth]{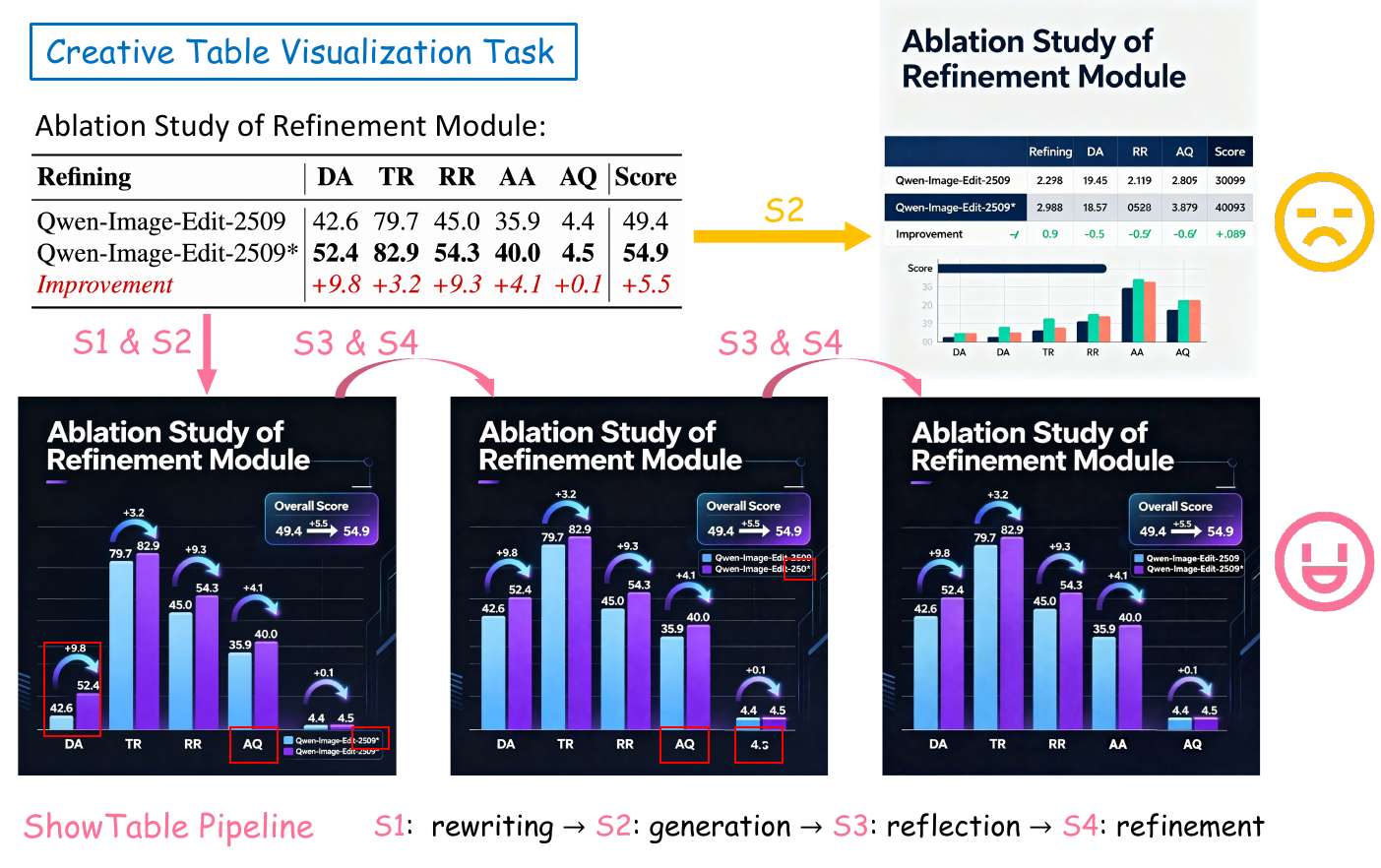}
    \caption{The illustration of our proposed creative table visualization task and ShowTable pipeline. Given a table about a specific topic, our task requires the model to produce a visualization infographic that is aesthetic and faithful to the data points. The ShowTable pipeline employs four steps: rewriting, generation, reflection, and refinement, thus achieving high-fidelity visualization. We use Wan2.5-Preview~\cite{wan2.5} here for generation and refinement.}
    \label{fig:intro}
\end{figure}

Image generation models have shown significant growth in quality in recent years~\cite{sd, sd3, dalle3, flux, t2i_r1, wei2025routing}. Recent advancements, particularly in unified models~\cite{januspro, showo, emu3, blip3o, bagel, uniworld, omnigen2, tang2025unilip,qwenimage, wan2.5}, have leveraged MLLMs for fine-grained understanding of complex multi-modal instructions~\cite{liu2025hybrid}, leading to more semantically coherent synthesis in general-purpose image generation and editing. Despite this rapid progress in the general domain, research is increasingly shifting towards more complex scenarios such as graphic design (\eg, poster generation~\cite{autoposter, postermaker, dreamposter} and text rendering~\cite{tuo2024anytext2, liu2024glyph, chen2024textdiffuser}). These tasks present a higher challenge as they demand not photorealism, but precise textual alignment and adherence to aesthetic principles of graphics.

However, within this structured domain, data-centric visualizations (\eg, charts and graphs derived from tables) present an even more formidable challenge. These tasks require not only understanding and reasoning of data but also a rigorous, high-fidelity mapping of quantitative data, where errors in visual proportions (e.g., bar heights, pie chart angles) or data labels render the output useless. To push beyond this limitation from unified reasoning and synthesis, we introduce a new and challenging task for unified models: \textbf{creative table visualization}, requiring a model to generate an infographic that is \textbf{aesthetic} and \textbf{faithful} to the data points within a given table. This task is defined by its dual requirements: 1) sophisticated reasoning for graphic design and aesthetic layout, and 2) strict, high-fidelity data-to-visual mapping from a source table. The ability to perform such synthesis could significantly streamline workflows in poster design, automatic slide generation, scientific communication, and other data-driven reporting tasks.

As shown in~\Cref{fig:intro}, current models still struggle with directly completing this task, as various errors about layout and data mapping may occur. Therefore, we propose \textbf{ShowTable}, a novel pipeline that synergizes MLLMs and diffusion models in a progressive, self-correcting loop, better addressing the task. As illustrated in~\Cref{fig:intro}, this process is inherently suited for achieving faithful data-to-image mapping, since the self-correcting loop iteratively refine initial inaccuracies. The framework relies on two core components: an MLLM acting as a central orchestrator and a diffusion model as the executor. The MLLM performs two key roles: 1) \textbf{Rewriting}, where it first reasons over the tabular data to plan an aesthetic visual sketch and rewrites the user prompt accordingly; and 2) \textbf{Reflection}, where it assesses the generated output and provides precise editing instructions. Correspondingly, the diffusion model executes two stages: 1) \textbf{Generation}, creating an initial figure based on the MLLM's sketch; and 2) \textbf{Refinement}, editing the figure based on the MLLM's reflective feedback.

To optimize the effectiveness of the pipeline, we conduct a pioneering exploration into training specific rewriting and refining capabilities. First, as~\Cref{fig:rewrite} illustrates, the rewriting module undertakes the crucial responsibility of reasoning and planning, its quality thus largely dictates the final outcome. Therefore, we produce 30K supervised-finetuning (SFT) data by captioning collected table visualizations and train this module. Second, as shown in~\Cref{fig:why_edit}, incapable refining models may degrade performance. This observation confirms that the refinement module could be a bottleneck in the pipeline.
Therefore, we first train a reward model (RM) using our constructed 30K comprehensive preference pairs. We then leverage this trained RM to optimize the diffusion model's refinement capability using reinforcement learning (RL) on a filtered set of 5K samples.

Furthermore, to evaluate the performance of models on this task, we introduce a comprehensive benchmark called \textbf{TableVisBench}. The benchmark features 800 challenging table instances, which are meticulously selected, filtered, labeled, and manually verified to ensure high quality. We design five key evaluation dimensions: data accuracy, text rendering, relative relationship, additional information accuracy, and aesthetic quality, comprehensively assessing the performance of different baselines.

Extensive experiments have demonstrated that our ShowTable pipeline significantly boosts the performance of all base models on TableVisBench. In conclusion, our contribution can be summarized as:
1) \textbf{New task}: We propose the creative table visualization task, which requires detailed reasoning and precise alignment ability, to challenge existing unified models. 
2) \textbf{Data}: We construct three automatic data construction pipelines to produce data for training, and present a \textit{TabelVisBench} benchmark, comprehensively evaluating the models' capability for creative table visualization task.
3) \textbf{Method}: We design a progressive self-correcting pipeline \textit{ShowTable} that cooperate an MLLM as central orchestrator with a diffusion model as executor. Experiments demonstrate the effectiveness of our framework.

\begin{figure*}[t]
    \centering
    \includegraphics[width=0.93\textwidth]{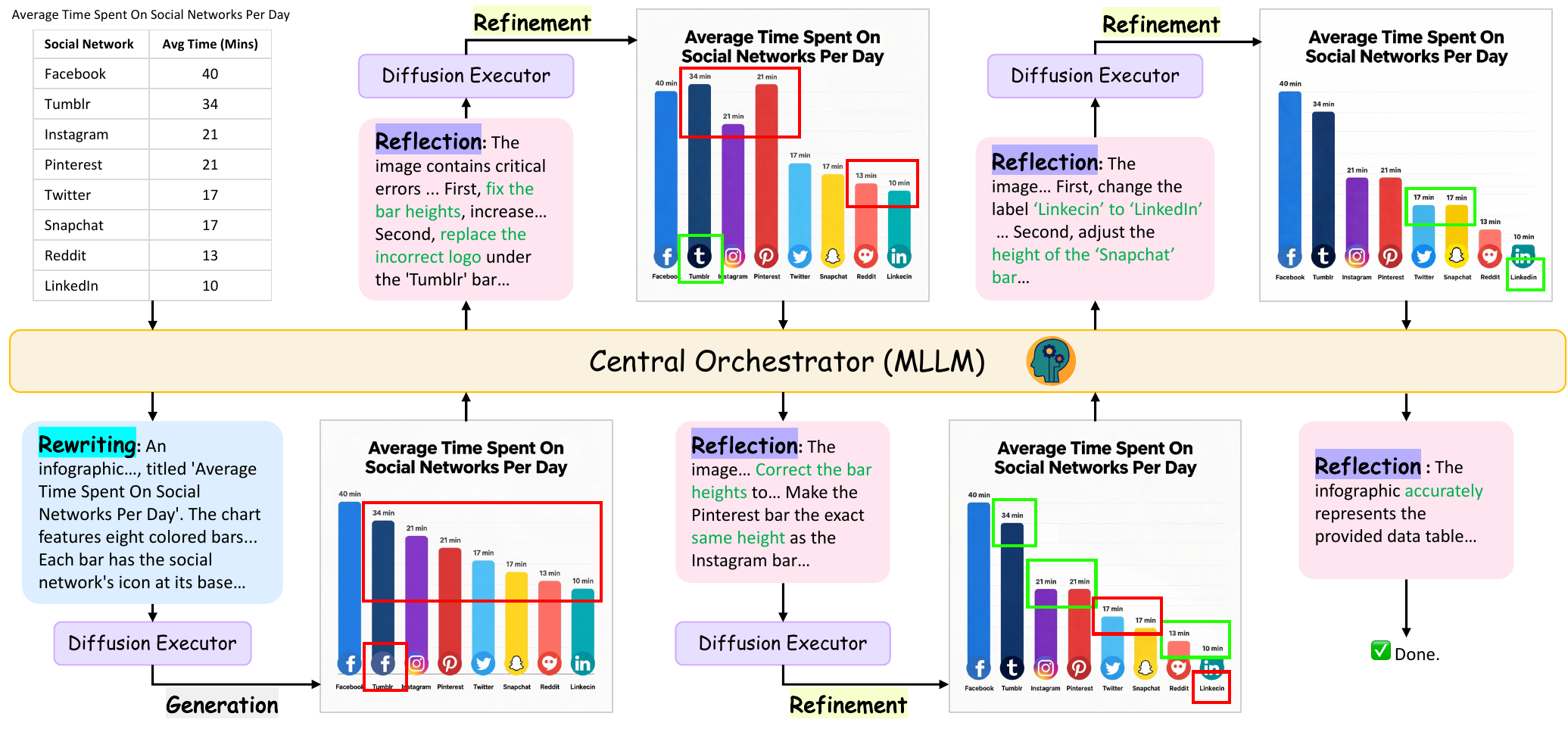}
    \caption{The proposed ShowTable pipeline, which synergizes an MLLM as the central orchestrator with a diffusion model as the executor. Given a table, the MLLM first rewrites a detailed prompt for the diffusion model's initial generation. The MLLM then iteratively reflects on the output to identify errors (marked in red) and provides precise instructions for refinement (corrected results shown in green).}
    \label{fig:structure}
\end{figure*}

\section{Related Work}
\label{sec:related_work}
\noindent\textbf{Image generation for graphic design.}
Recent advances in image generation have significantly enhanced the quality and semantic controllability of synthesized content in general-purpose scenarios~\cite{sd, sd3, dalle3, flux, t2i_r1, januspro, showo, emu3, blip3o, bagel, uniworld, omnigen2, qwenimage, wan2.5}. This progress has spurred growing interest in more complex generation tasks such as graphic design. AnyText~\cite{tuo2023anytext, tuo2024anytext2}, Glyph-ByT5~\cite{liu2024glyph}, and TextDiffuser~\cite{chen2024textdiffuser} have made significant efforts in text rendering, focusing on accurate textual element incorporation. AutoPoster~\cite{autoposter}, PosterMaker~\cite{postermaker}, and DreamPoster~\cite{dreamposter} focus on poster generation, which addresses aesthetic layout planning. The recently proposed Qwen-Image~\cite{qwenimage} also emphasizes capabilities in complex text rendering and infographic design. Compared to these scenarios, our proposed creative table visualization task presents a more formidable challenge. Our task demands not only reasoning about graphic design for aesthetic layout but also a precise data-to-visual mapping for faithful representation of tabular content.

\noindent\textbf{Reasoning and reflection paradigm of MLLMs.}
The reasoning and reflection abilities of MLLMs are gaining increasing attention for both understanding and generation. Some studies leverage the reasoning capability of MLLMs to achieve better image understanding, often termed \textit{thinking with images}~\cite{thinkingwithgeneratedimages, zheng2025deepeyes, zhuo2024lumina, Visualsketchpad,bao2024cores}. For image generation, some works~\cite{bagel, t2i_r1, fang2025got, xiao2025mindomni} introduce a text-based reasoning step, enhancing image generation performance. Recently, emerging research has begun to leverage the MLLM reflection process to refine the image generation itself, aiming to enhance instruction-following in complex, general-purpose scenes~\cite{IRG, huang2024dialoggen, wang2024genartist}. However, this category of research remains focused on general domains. The high information density and specific structural constraints of our task pose a greater challenge for MLLMs with combining reasoning, planning, and reflection abilities.

\section{Method}
\label{sec:method}

\subsection{ShowTable Pipeline}
\noindent\textbf{Overview.}
To address the challenge of creative table visualization and push model reasoning in the infographic domains, we propose a base pipeline, \textbf{ShowTable}. To confront the limitations of existing models in preserving visual relational reasoning and information consistency, ShowTable introduces a structured \textit{Rewriting $\rightarrow$ Generation $\rightarrow$ Reflection $\rightarrow$ Refinement} workflow. This iterative self-correction loop is specifically designed to address the challenges of creative table visualization, where standard generation models often fail, producing logical inconsistencies or text rendering errors. As shown in~\Cref{fig:structure}, ShowTable progressively improves the quality and detail fidelity by repeated reflection and refinement.

\noindent\textbf{Rewriting.}
There exists a significant distinction between raw tables and typical image generation prompts. Table inputs (\eg, in markdown format) possess high information density, with each data point encapsulating complex relational semantics. Visualizing such dense and non-redundant data necessitates deliberate reasoning about data presentation and layout. When a markdown-formatted table is directly used as a prompt, the generation model tends to misinterpret the task, often trying to render the table itself rather than visualizing its underlying data, as illustrated in~\Cref{fig:rewrite}. To bridge this gap, we use an MLLM to perform semantic and structural reasoning with compositional planning. It translates the data-dense table into a detailed descriptive prompt that generation models can execute.

\noindent\textbf{Generation.}
Upon obtaining the rewritten prompt, we feed it into a pre-trained text-to-image (T2I) model to produce an initial image. As shown in~\Cref{fig:structure}, this preliminary result generally captures the overall layout and aesthetic. However, it often contains critical errors in high-fidelity details, such as incorrect data-to-visual correspondence (e.g., bar heights), axis label misalignments, or suboptimal text rendering. This imperfect initial generation serves as the foundation for the subsequent reflection and refinement stages.

\noindent\textbf{Reflection.}
Given the strict fidelity required for table visualization, we employ a reflection module. This module uses an MLLM to perform a critical audit of the generated image. By cross-referencing the original table with the generated image, the MLLM identifies inconsistencies and inaccuracies. It then formulates a set of precise, actionable editing instructions to correct these errors.

\noindent\textbf{Refinement.}
Finally, as shown in~\Cref{fig:structure}, these edit instructions are provided to an image editing model, which executes the corrections on the initial image to produce the refined, high-fidelity visualization.

\subsection{Training Details}
\label{sec:training_detail}
\begin{figure}[!t]
    \centering
    \includegraphics[width=0.42\textwidth]{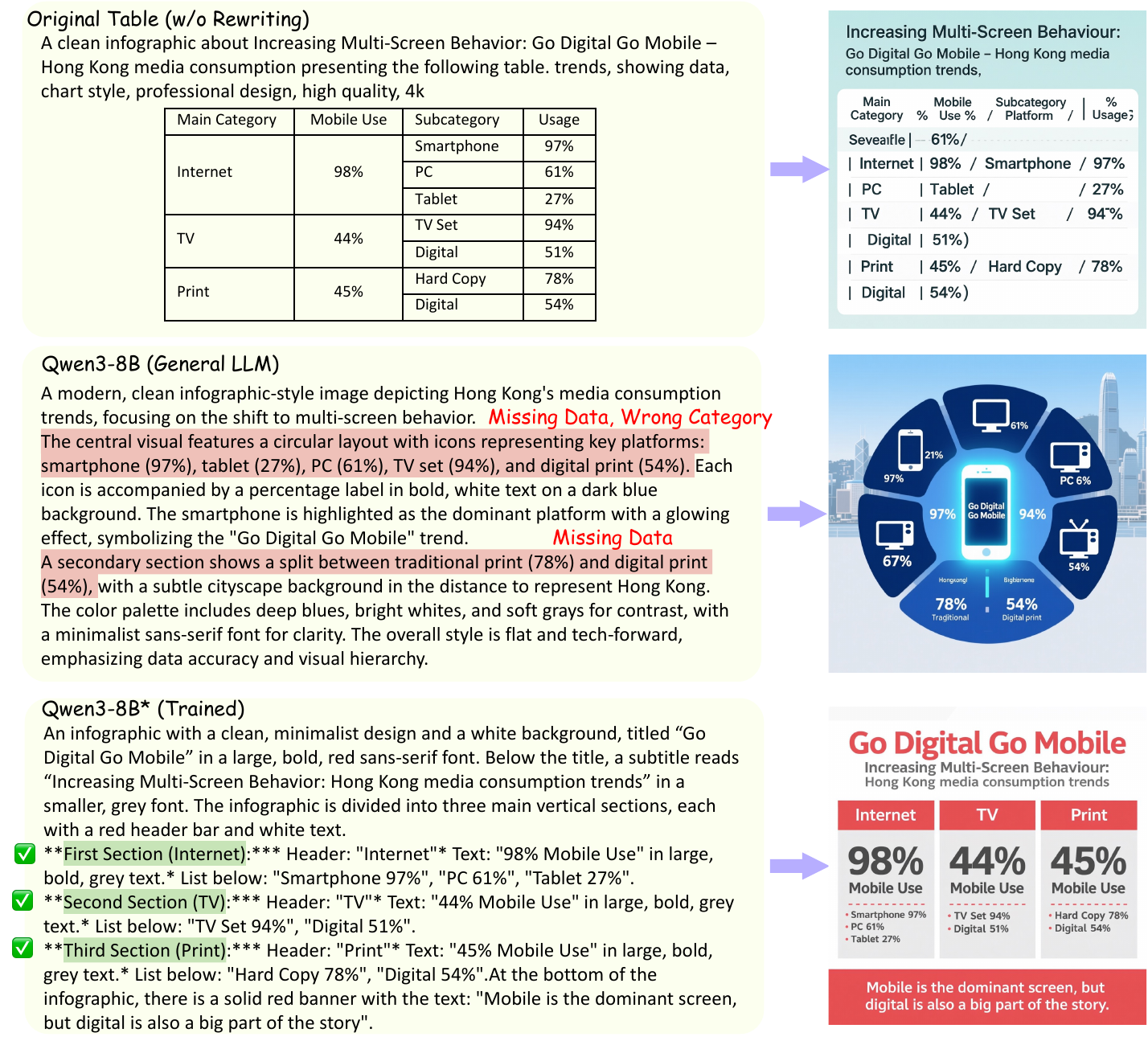}
    \caption{Generation comparison among different rewriting. The model tends to directly render the table text, failed to visualize when disabling rewriting. General LLMs also tend to miss some data or wrongly classify data, especially for complex tables.}
    \label{fig:rewrite}
\end{figure}

\noindent\textbf{Rewriting module.}
Although general LLMs have demonstrated powerful reasoning abilities on various tasks, their performance on table reasoning and compositional planning can be further enhanced. As~\Cref{fig:rewrite} shows, they may miss data points or plan poorly when encountering complex data (e.g., multi-layered structures). As this is the critical step responsible for basic layout design and content planning, fine-tuning the rewrite module is expected to significantly enhance the overall performance. Therefore, we fine-tune a specific rewrite module from Qwen3-8B~\cite{qwen3} to address this challenge. With our constructed 30K rewriting training data (\Cref{sec:train_data}), we train the module with the standard next-token prediction format:
\begin{equation}
\mathcal{L}_\text{rewrite}= -\frac{1}{N} \sum_{n=1}^N \log(\hat{y}_{n, k_n})
\end{equation}
where $N$ denotes the sequence length, and $\hat{y}_{n, k_n}$ is the predicted probability for the true class $k_n$ at position $n$.

\begin{figure}[!t]
    \centering
    \includegraphics[width=0.45\textwidth]{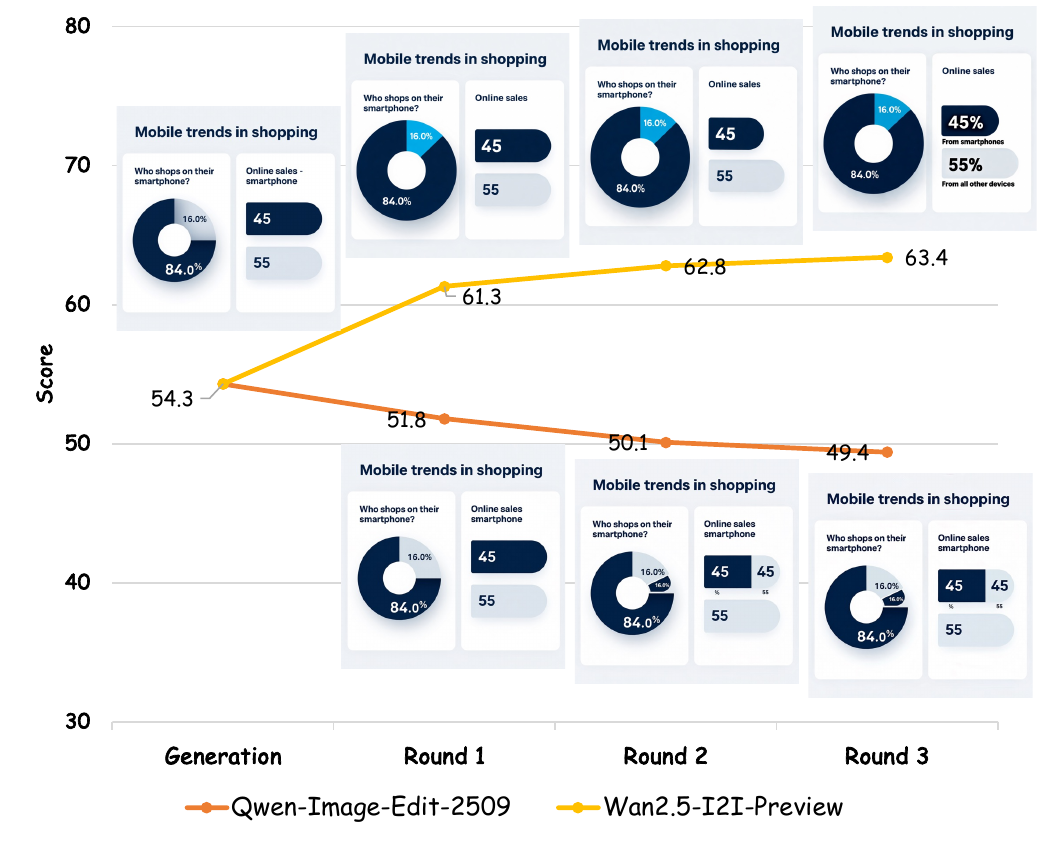}
    \caption{Demonstrating the refinement model's capability as a critical pipeline bottleneck. The base model's performance degrades with each correction, indicating an inability to process iterative feedback. In contrast, the Wan2.5-I2I-Preview shows consistent improvement. This confirms our pipeline structure is sound and that the bottleneck is the model's capability, motivating our specialized training for the refinement module.}
    \label{fig:why_edit}
\end{figure}

\noindent\textbf{Refinement module.}
The refinement stage is critical for correcting errors, but it may also present a significant potential bottleneck. As illustrated in~\Cref{fig:why_edit}, our preliminary experiments revealed this challenge. When employing a recent SOTA Qwen-Image-Edit~\cite{qwenimage} within our iterative loop, we observed a performance degradation with each correction round. This raised a critical question: is the pipeline's self-correction logic flawed, or is the editing model's capability insufficient? To investigate this variable, we introduced Wan2.5-I2I-Preview~\cite{wan2.5}, another editing model known for fine-grained controllability, and found that performance indeed increased with each iteration. This indicates that our pipeline structure is sound, but its effectiveness is fundamentally constrained by the refinement model's ability to execute precise edits. Therefore, to resolve this bottleneck, we train the refinement module using RL, specifically employing the Group Relative Policy Optimization (GRPO) algorithm~\cite{deepseekr1}. This approach requires an accurate reward signal. However, evaluating rendering quality is complex, integrating multiple dimensions. As recent pretrained MLLMs struggle to directly provide consistent, accurate scalar scores for such assessments~\cite{luo2025editscore} (see Appendix), there remains a need to develop a specialized reward model (RM). Using our constructed 30K pairwise preference dataset (\Cref{sec:train_data}), we fine-tune a Qwen2.5-VL-3B~\cite{qwen2.5vl} model, $f_\theta$, as our quality assessor. The model is trained to distinguish positive ($x_w$) and negative ($x_l$) samples for a given prompt $p$ using the Bradley-Terry loss~\cite{bradley1952bradterry}:
\begin{equation}
    \mathcal{L}_{BT} = -\mathbb{E} \left[ \log \sigma \left( f_\theta(x_w, p) - f_\theta(x_l, p) \right) \right],
\end{equation}
where $\sigma(\cdot)$ is the Sigmoid function. To enhance training efficiency, the scalar reward score is computed by extracting the probabilities corresponding to digits 0–9 from the output logits and averaging their sum. Finally, using our constructed 5K refinement data (\Cref{sec:train_data}), we follow Flow-GRPO~\cite{flowgrpo} to perform the RL. The full reward signal combines our trained $f_\theta$ with an existing aesthetic reward model, ImageReward~\cite{imagereward}, to optimize the training.

\begin{figure*}[t]
    \centering
    \includegraphics[width=0.95\textwidth]{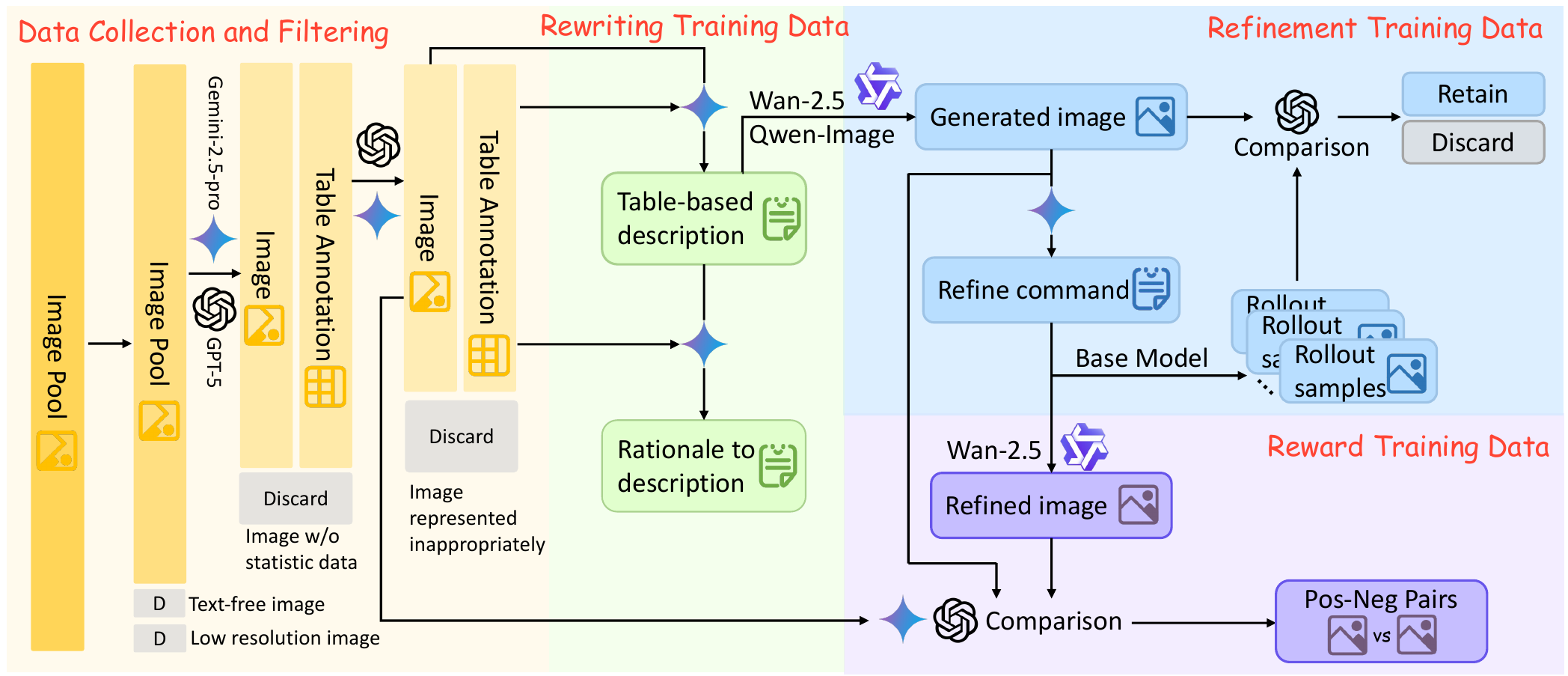}
    \caption{Dataset construction pipeline. We initially collect and filter images from public datasets with SOTA MLLMs, and then propose three different kinds of training data construction pipelines: rewriting training data, refinement training data, and reward training data.}
    \label{fig:data_construct}
\end{figure*}

\section{Dataset and Benchmark}
\subsection{Data Collection and Filtering}
\label{sec:data_collect}
To support both training and benchmarking for our proposed task, we construct a dataset of 30K high-quality table-image pairs and an additional benchmark of 800 evaluation samples. We initiate this by collecting raw images from diverse public datasets, including SlideVQA~\cite{tanaka2023slidevqa}, OpenImages~\cite{kuznetsova2020open}, and Cambrian-10M~\cite{tong2024cambrian}. We then applied the rigorous filtering and annotation pipeline shown in \Cref{fig:data_construct}. First, we discard low-resolution images (under $200\times200$) and images lacking text, as identified by PaddleOCR~\cite{cui2025paddleocr}. Following this, we utilize SOTA MLLMs (i.e., Gemini-2.5-pro\cite{gemini2.5pro} and GPT-5~\cite{gpt5}) to perform a crucial filtering and annotation step: the models filter out images that do not feature statistical data as the main body, and simultaneously annotate the table information in markdown format for all remaining images. To ensure the quality of these annotations, we implemented a consensus-based verification process. Both MLLMs independently annotate the filtered images. We then retained only the samples for which the annotations from both models were consistent and mutually approved, resulting in our final set of 30K high-quality table-image pairs.

\subsection{Training Data Construction}
\label{sec:train_data}
Based on the collected 30K table-image pairs, we propose three automated training data construction pipelines to support the distinct stages of our method.

\noindent\textbf{Rewriting training data.}
As the rewriting module is critical for semantic reasoning and compositional planning, high-quality SFT data is essential. As shown in~\Cref{fig:data_construct}, we first prompt Gemini-2.5-pro to generate a detailed description of the ground-truth image based on the annotated table, covering data points, layout, color, and background. We then prompt Gemini-2.5-pro again, providing both the table and the new description, to generate a chain-of-thought rationale that explains the conversion process. This results in 30K data pairs (\{table, rationale\} $\rightarrow$ \{description\}), which are used to fine-tune the rewriting module.

\noindent\textbf{Refinement training data.}
To generate data for RL, we first use the descriptions from the previous step to generate initial images using Wan2.5-t2i-preview~\cite{wan2.5} and Qwen-Image~\cite{qwenimage}. We then use Gemini-2.5-pro to audit these generated images and produce corresponding refinement instructions. We apply a rigorous filtering strategy to ensure RL training stability. For each sample (initial image + instruction), we instruct our base editing model (Qwen-Image-Edit~\cite{qwenimage}) to generate five refined candidates. These candidates are compared against the initial image by a powerful MLLM assessor (GPT-5). We discard samples where all five attempts are judged as worse, or all five as better, than the original. This filtering isolates samples that are too hard or too easy for base model, yielding 5K challenging samples ideal for refinement training.

\noindent\textbf{Reward training data.}
Our RL approach requires a reliable RM. Given that MLLMs are unstable in providing direct point-wise scores~\cite{luo2025editscore}, we construct a preference dataset for converting the preference into a specific score assessor. We use GPT-5 and Gemini-2.5-pro to compare each image pair, followed by voting, and finally generate 30K positive-negative image pairs for training the RM, as shown in~\Cref{fig:data_construct}. These pairs are sourced from three comparisons: 1) images refined by Wan2.5 versus images generated by Wan2.5 or Qwen; 2) images generated by Wan2.5 versus those by Qwen; and 3) collected ground-truth images versus generated or refined images.

\begin{table}[!t]
\setlength\tabcolsep{3pt}
\centering
\caption{Performance comparison of recent strong open-sourced baselines on our TableVisBench. ``RW'' refers to the rewriting module, ``REF'' refers to the reflection and refinement process. We mark the improvement of our proposed pipeline for each base generation model.}
\label{tab:main}
\resizebox{0.48\textwidth}{!}{
\begin{tabular}{l|ccccc|c}
\toprule
\textbf{Methods}& \textbf{DA} & \textbf{TR}&\textbf{RR}&\textbf{AA}&\textbf{AQ}&\textbf{Score} \\
\midrule
\textit{Reference Image}&\textit{97.7}&\textit{99.5}&\textit{86.4}&\textit{96.6}&\textit{4.2}&\textit{84.4}\\
\midrule
Flux~\cite{flux} &12.1&46.7&28.9&18.7&4.0&29.3 \\
RW+Flux&12.0&52.3&27.0&25.3&4.4&32.1 \\
RW+Flux+REF&20.3&63.1&31.8&24.0&4.3&36.4 \\
\textit{\darkred{Improvement}}&
\textit{\darkred +8.2}&\textit{\darkred +16.4}&\textit{\darkred +2.9}&\textit{\darkred +5.3}&\textit{\darkred +0.3}&\textit{\darkred +7.1}\\
\midrule
Bagel~\cite{bagel}&0.1&1.6&14.2&7.7&2.7&10.1 \\
RW+Bagel&3.4&18.3&28.9&13.0&3.4&19.5 \\
RW+Bagel+REF&18.3&54.8&36.7&15.9&3.8&32.7 \\
\textit{\darkred Improvement}&
\textit{\darkred +18.2}&\textit{\darkred +53.2}&\textit{\darkred +22.5}&\textit{\darkred +8.2}&\textit{\darkred +1.1}&\textit{\darkred +22.6}\\
\midrule
Blip3o-Next~\cite{chen2025blip3onext}&0.4&18.0&4.4&6.2&2.5&10.8 \\
RW+Blip3o-Next&0.5&14.5&19.1&7.6&2.9&14.1 \\
RW+Blip3o-Next+REF&21.3&63.9&33.4&19.2&3.6&34.8 \\
\textit{\darkred Improvement}&
\textit{\darkred +20.9}&\textit{\darkred +45.9}&\textit{\darkred +29.0}&\textit{\darkred +13.0}&\textit{\darkred +1.1}&\textit{\darkred +24.0}\\
\midrule
UniWorld-V1~\cite{uniworld}&3.0&18.3&14.7&2.9&3.5&14.8 \\
RW+UniWorld-V1&4.0&20.8&23.7&11.6&3.3&18.6 \\
RW+UniWorld-V1+REF&18.7&54.6&37.6&18.8&3.8&33.5 \\
\textit{\darkred Improvement}&
\textit{\darkred +15.7}&\textit{\darkred +36.3}&\textit{\darkred +22.9}&\textit{\darkred +15.9}&\textit{\darkred +0.3}&\textit{\darkred +18.7}\\
\midrule
OmniGen2~\cite{omnigen2}&3.1&17.8&13.5&2.6&3.5&14.4 \\
RW+OmniGen2&4.0&32.1&25.0&9.5&3.9&21.9 \\
RW+OmniGen2+REF&16.2&49.8&30.6&13.8&3.9&29.9 \\
\textit{\darkred Improvement}&
\textit{\darkred +13.1}&\textit{\darkred +32.0}&\textit{\darkred +17.1}&\textit{\darkred +11.2}&\textit{\darkred +0.4}&\textit{\darkred +15.5}\\
\midrule
Qwen-Image~\cite{qwenimage}&47.5&90.9&26.1&14.1&4.3&44.3\\
RW+Qwen-Image&51.2&83.1&50.1&40.9&4.6&54.3 \\
RW+Qwen-Image+REF&52.4&82.9&54.3&40.0&4.5&54.9 \\
\textit{\darkred Improvement}&
\textit{\darkred +4.9}&\textit{\darkred -8.0}&\textit{\darkred +28.2}&\textit{\darkred +25.9}&\textit{\darkred +0.2}&\textit{\darkred +10.6}\\
\bottomrule
\end{tabular}%
}
\end{table}

\subsection{Benchmark Design}
To accurately evaluate performance on our creative table visualization task, we construct \textbf{TableVisBench}, a benchmark containing 800 challenging table-based instances. The collection and filtering process is similar to that of the training data, but with an additional step of manual verification and correction for all samples. Detailed statistics about the benchmark are provided in the Appendix.

For comprehensive benchmarking, we conduct multi-view evaluation~\cite{liu2025capability} with five well-designed dimensions, focusing not only on factual accuracy but also on logical coherence and visual aesthetics. Instead of using an MLLM as a subjective scorer, we leverage it as a quality assurance analyst. For the first four dimensions, the MLLM is prompted to identify and quantify specific errors within the generated chart. The final score is then deterministically calculated based on the number of reported errors, thereby mitigating the instability and bias associated with direct LLM-based scoring. The final dimension is quantitatively assessed using a dedicated aesthetic scoring model. The five dimensions are as follows:

\noindent\textbf{Data Accuracy (DA).}
This dimension verifies that every single data point from the source table is accurately represented in the generated image, ensuring none are missing, incorrect, or simply rendered as raw table text.

\noindent\textbf{Text Rendering (TR).}
This dimension focuses on the legibility and correctness of all textual elements in the image.

\noindent\textbf{Relative Relationship (RR).}
This dimension assesses the core visualization logic, \ie, whether the visual proportions of chart elements (e.g., bar heights, slice angles) correctly reflect the quantitative relationships between data points.

\noindent\textbf{Additional information Accuracy (AA).}
This dimension inspects the accuracy and appropriateness of contextual information added by the model (not present in the source table), such as axes, ticks, gridlines, and extraneous artifacts.

\noindent\textbf{Aesthetic Quality (AQ).}
Independent of factual correctness, this dimension evaluates the overall visual appeal of the generated chart, including its layout, color palette, typography, and design creativity.

The scores for the first four dimensions (DA, TR, RR, AA) range from 0 to 100, while AQ ranges from 0 to 10. We calculate the final score by:
\begin{equation}
    \text{Score} = (\text{DA} + \text{TR} + \text{RR} + \text{AA} + 10\times\text{AQ}) / 5
\end{equation}
We provide detailed information about these dimensions in the Appendix. To validate the reliability of our benchmark, we test our dimensions on the collected ground-truth images. As shown in~\Cref{tab:main}, these high-quality images achieve very high scores, confirming that our evaluation metrics are well-aligned with human-annotated quality.

\begin{table}[!t]
\setlength\tabcolsep{4pt}
\centering
\caption{Ablation of the rewriting module of our ShowTable. All results reported in the table are the generated results after the rewriting process without reflection and refinement. The generation module utilized is Qwen-Image (8 steps distilled).}
\label{tab:rewrite}
\resizebox{0.47\textwidth}{!}{
\begin{tabular}{l|ccccc|c}
\toprule
\textbf{Rewrite} & \textbf{DA} & \textbf{TR}&\textbf{RR}&\textbf{AA}&\textbf{AQ}&\textbf{Score} \\
\midrule
False &47.5&90.9&26.1&14.1&4.3&44.3\\
Qwen3-8B~\cite{qwen3} &30.6&71.5&46.6&34.1&5.1&46.8\\
GPT-5~\cite{gpt5} &35.9&78.5&47.8&41.8&5.2&51.2\\
Gemini-2.5-pro~\cite{gemini2.5pro} &40.8&79.9&53.9&41.1&5.1&53.3\\
\rowcolor[HTML]{cbcbcb} 
\textbf{Qwen3-8B*}&51.2&83.1&50.1&40.9&4.6&54.3\\
\midrule
Reference-Caption &50.3&83.4&55.1&42.8&4.5&55.3\\
\bottomrule
\end{tabular}%
}
\end{table}

\section{Experiments}
\label{sec:exp}
\subsection{Implementation Setup}
The modular design of our ShowTable pipeline allows for flexible combinations of MLLMs and diffusion models. In our default configuration, the rewriting module is trained based on Qwen3-8B~\cite{qwen3}, and the refinement module is trained based on a distilled 8-step version~\cite{qwenimagelightning} of Qwen-Image-Edit-2509~\cite{qwenimage} to accelerate RL training. For the reflection module, we employ GPT-5-2025-08-07~\cite{gpt5}. For fair comparison, all baselines that use Qwen-Image for generation or Qwen-Image-Edit-2509 for refinement also use these same distilled versions~\cite{qwenimagelightning}. We set the maximum self-correction round to 3, though the process can terminate early if the reflection module deems an image satisfactory. More detailed settings can be found in the Appendix.

\subsection{Main Results}
We comprehensively evaluate the effectiveness of the ShowTable pipeline by applying it to several advanced T2I generation models, including Flux~\cite{flux}, Bagel~\cite{bagel}, Blip3o-Next~\cite{chen2025blip3onext}, UniWorld-V1~\cite{uniworld}, OmniGen2~\cite{omnigen2}, and Qwen-Image~\cite{qwenimage}. The evaluation is conducted on our TableVisBench based on the five dimensions. We systematically compare performance under three configurations: 1) The base generation model alone (Base); 2) The base model prefixed with our rewriting module (RW+Base); 3) The full pipeline, integrating all modules (RW+Base+REF).

\noindent\textbf{Quantitative analysis.}
The results, presented in~\Cref{tab:main}, demonstrate three clear findings. First, base models alone are incapable of our proposed challenging task. Some models, such as Bagel and Blip3o-Next, score near zero (0.1 and 0.4, respectively) on Data accuracy, indicating a fundamental failure to translate table data into visual components. Second, the rewriting (RW) module is critical for reasoning and planning. Simply adding the RW module significantly boosts performance, especially in logical coherence. For instance, with Qwen-Image, the RR score jumps from 26.1 to 50.1. This shows that converting raw markdown into a reasoned, descriptive prompt is an essential first step. Third, the reflection and refinement (REF) loop is essential for accuracy. The full pipeline (RW+Base+REF) achieves the best overall score in all cases. This step yields the most significant gains in correctness-based metrics. With Blip3o-Next, the full pipeline improves DA from 0.5 to 21.3 and TR from 14.5 to 63.9. Moreover, our approach also shows strong adaptability. For models with weaker baseline capabilities (\eg, Bagel), our pipeline provides a substantial boost, improving the final score by +22.6 points, respectively. For strong base models like Qwen-Image, the full pipeline unlocks their potential, achieving the highest scores in key metrics and demonstrating a powerful synergistic effect, raising the overall score from 44.3 to 54.9.

\begin{table}[!t]
\setlength\tabcolsep{4pt}
\centering
\caption{Ablation of the different models of the reflection module. We use our trained Qwen3-8B as the rewriting module, Qwen-Image (distilled) as the generation module, and Qwen-Image-Edit-2509 (distilled) / our trained model as the refining module here.}
\label{tab:reflection}
\resizebox{0.47\textwidth}{!}{
\begin{tabular}{l|ccccc|c}
\toprule
\textbf{Reflection} & \textbf{DA} & \textbf{TR}&\textbf{RR}&\textbf{AA}&\textbf{AQ}&\textbf{Score} \\
\midrule
\multicolumn{7}{l}{\textit{Refinement: Qwen-Image-Edit-2509}}\\
Qwen3-VL-235B~\cite{qwen3vl} &37.7&78.1&45.1&31.3&\textbf{4.4}&47.2\\
Gemini-2.5-pro~\cite{gemini2.5pro} &\textbf{43.2}&\textbf{81.2}&\textbf{46.6}&\textbf{41.1}&\textbf{4.4}&\textbf{51.2} \\
GPT-5~\cite{gpt5} &{42.6}&79.7&45.0&35.9&\textbf{4.4}&49.4\\
\midrule
\multicolumn{7}{l}{\textit{Refinement: Qwen-Image-Edit-2509* (trained by ours)}}\\
Qwen3-VL-235B~\cite{qwen3vl} &46.9&81.5&48.3&37.1&\textbf{4.5}&51.8\\
Gemini-2.5-pro~\cite{gemini2.5pro} &48.1&\textbf{83.0}&48.8&\textbf{44.8}&\textbf{4.5}&53.9 \\
GPT-5 ~\cite{gpt5}&\textbf{52.4}&82.9&\textbf{54.3}&40.0&\textbf{4.5}&\textbf{54.9}\\
\bottomrule
\end{tabular}%
}
\end{table}

\begin{table}[!t]
\setlength\tabcolsep{2pt}
\centering
\caption{Ablation of different models for the refinement module. We use our trained Qwen3-8B as the rewriting module, Qwen-Image (distilled) as the generation module, and GPT-5 as the reflection module here. * donates the method trained by ours.
}
\label{tab:refine}
\resizebox{0.47\textwidth}{!}{
\begin{tabular}{l|ccccc|c}
\toprule
\textbf{Refining} & \textbf{DA} & \textbf{TR}&\textbf{RR}&\textbf{AA}&\textbf{AQ}&\textbf{Score} \\
\midrule
Qwen-Image-Edit-2509 &42.6&79.7&45.0&35.9&4.4&49.4\\
Qwen-Image-Edit-2509*&\textbf{52.4}&\textbf{82.9}&\textbf{54.3}&\textbf{40.0}&\textbf{4.5}&\textbf{54.9} \\ 
{\darkred \textit{Improvement}}&\textit{\darkred +9.8}&\textit{\darkred +3.2}&\textit{\darkred +9.3}&\textit{\darkred +4.1}&\textit{\darkred +0.1}&\textit{\darkred +5.5}\\
\midrule
Wan2.5-I2I-Preview &64.2&84.7&64.6&59.7&4.4&63.4\\
\bottomrule
\end{tabular}%
}
\end{table}

\begin{figure*}[!t]
    \centering
    \includegraphics[width=0.9\textwidth]{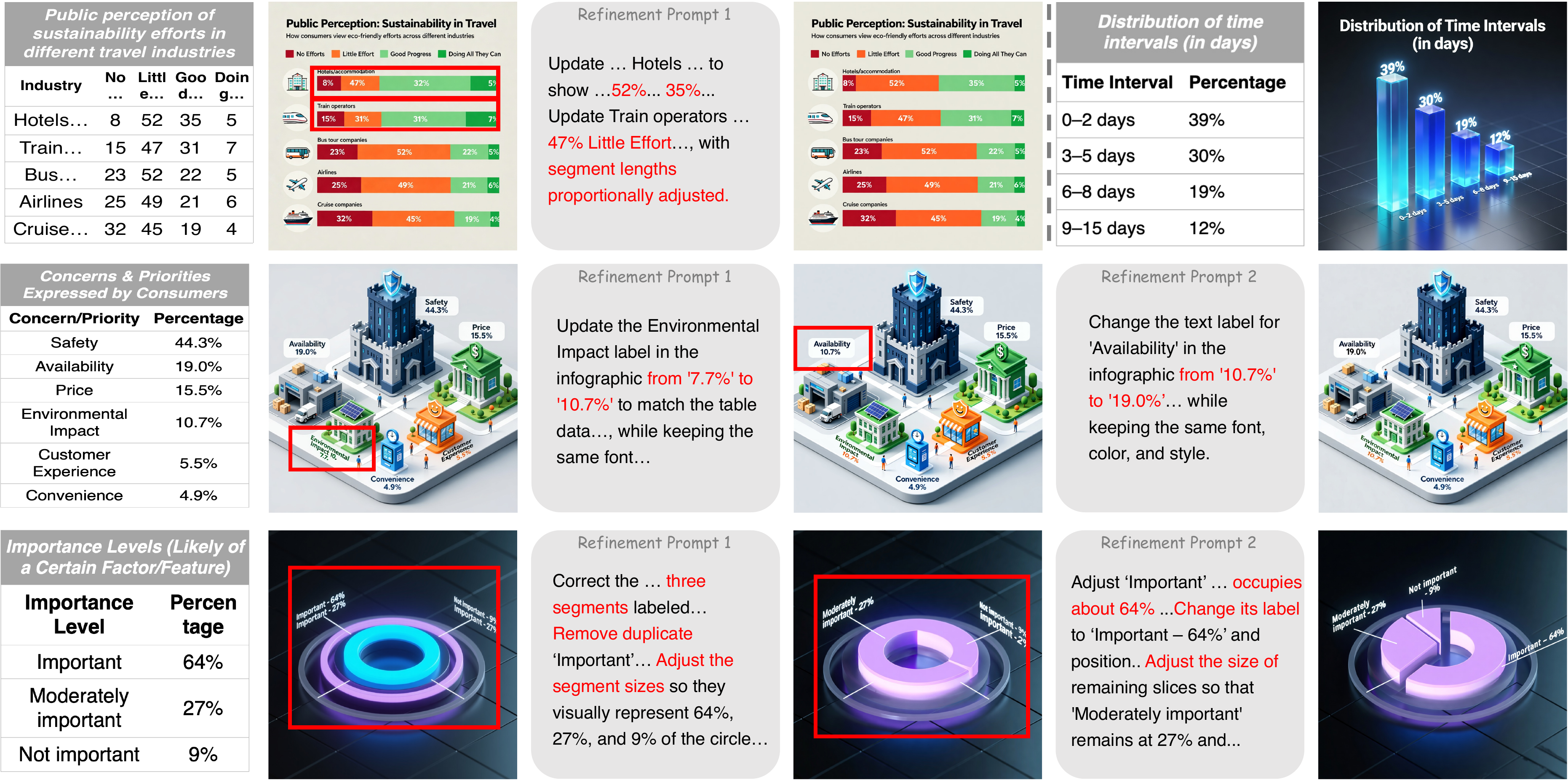}
    \caption{Case studies of the ShowTable pipeline with four examples: top row shows one case requiring one-round reflection and refinement (left) and one proper initial generation (right), while other rows display results refined through multiple reflection rounds. Examples demonstrate adaptive correction of text, proportions, and visual elements. We mark the error parts with red boxes for better reading.}
    \label{fig:case}
\end{figure*}

\noindent\textbf{Qualitative analysis.}
Qualitative results in~\Cref{fig:case} also demonstrate ShowTable's ability to produce creative yet accurate visualizations across diverse table structures. The reflection-refinement mechanism effectively corrects various error types: misrendered text and numbers (Row 1-Left, Row 2), incorrect proportional relationships (Row 1-Left), and improper visual element representations (Row 3). Cases requiring zero refinement (Row 1-Right) confirm the pipeline's adaptive efficiency. These results validate the robustness of ShowTable in achieving faithful and creative table-to-image translation.

\subsection{Ablation Studies}

\noindent\textbf{Rewriting module.}
We evaluate different rewriting strategies using the Qwen-Image generation model. As shown in~\Cref{tab:rewrite}, compared settings include: 1) no rewriting (False), 2) general-purpose LLMs (Qwen3-8B, GPT-5, Gemini-2.5-pro), 3) our fine-tuned model (Qwen3-8B*), and 4) an upper-bound using reference captions (Reference-Caption). All results are based on the initial generation (no refinement). Our fine-tuned module (Qwen3-8B*) achieves the highest overall score (54.3) among all evaluated rewriting methods, demonstrating a strong, well-balanced performance. Notably, it attains the best DA (51.2), confirming that specialized training is more effective at preserving data integrity than general-purpose LLMs (30.6-40.8). While all rewriting methods slightly reduce the TR score compared to the baseline (which often just renders the table text directly), our model's substantial gains in DA (+3.7 vs base) and RR (+24.0 vs base) justify this trade-off. Notably, our fine-tuned model even surpasses the Reference-Caption's DA (51.2 vs. 50.3), further underscoring the advantage of training. While Gemini-2.5-pro achieves the highest RR score, our model remains highly competitive and delivers the best overall performance, highlighting its robustness.

\noindent\textbf{Reflection module.}
We evaluate the effectiveness of different MLLMs as the reflection module. As shown in~\Cref{tab:reflection}, we use our fine-tuned rewriting module (Qwen3-8B*) and the Qwen-Image generation model. We then compare the final performance with Qwen3-VL-235B, Gemini-2.5-pro, and GPT-5 as reflection modules. The results demonstrate the significant impact of the reflection module on output quality. When using our trained refinement model (bottom half of table), GPT-5 achieves the strongest performance (54.9), excelling in Data Accuracy (52.4) and Relative Relationship (54.3). When paired with the base Qwen-Image-Edit module (top half of table), Gemini-2.5-pro delivers the best results (51.2). These findings confirm that more capable reflection models consistently enhance output quality, providing practical guidance for model selection.

\begin{table}[!t]
\setlength\tabcolsep{3pt}
\centering
\caption{Ablation of the maximum refining rounds. We keep the same setting as \Cref{tab:refine}, and compare results of each round.}
\label{tab:round}
\resizebox{0.47\textwidth}{!}{
\begin{tabular}{l|c|ccccc|c}
\toprule
\textbf{Methods}& \textbf{Num} & \textbf{DA} & \textbf{TR} & \textbf{RR} & \textbf{AA} & \textbf{AQ} & \textbf{Score} \\ \midrule
w/o refine& 0            & 51.2        & 83.1        & 50.1        & 40.9        & 4.6         & 54.3           \\ \midrule
\multirow{3}{*}{Qwen-Image-Edit-2509}& 1            & \textbf{45.6}        & \textbf{81.7}        & \textbf{48.3}        & \textbf{38.3}        & \textbf{4.5}         & \textbf{51.8}           \\
       & 2            & 42.6        & 80.7        & 46.2        & 36.2        & \textbf{4.5}         & 50.1           \\
       & 3            & 42.6        & 79.7        & 45.0        & 35.9        & 4.4         & 49.4           \\ \midrule
\multirow{3}{*}{Qwen-Image-Edit-2509*} & 1            & 50.0        & 82.6        & 51.3        & 39.4        & 4.5         & 53.7           \\
       & 2            & 50.4        & \textbf{83.0}        & 52.9        & \textbf{41.5}        & \textbf{4.6}         & 54.8           \\
       & 3            & \textbf{52.4}        & 82.9        & \textbf{54.3}        & 40.0        & 4.5         & \textbf{54.9}           \\ \midrule
\multirow{3}{*}{Wan2.5-I2I-Preview}    & 1            & 60.7&\textbf{85.2}&61.8&53.8&\textbf{4.5}&61.3           \\
       & 2            & 63.3&85.1&64.1&57.5&4.4&62.8           \\
       & 3            & \textbf{64.2}&84.7&\textbf{64.6}&\textbf{59.7}&4.4&\textbf{63.4}           \\ \bottomrule
\end{tabular}%
}
\end{table}

\noindent\textbf{Refinement module.}
We evaluate different refining modules under fixed rewriting (our fine-tuned Qwen3-8B*) and reflection (GPT-5) conditions, with results in \Cref{tab:refine}. Our trained refinement model (Qwen-Image-Edit-2509*) achieves significant improvements over its base model, increasing the overall score from 49.4 to 54.9 (+5.5). The model proves notable gains in DA (+9.8) and RR (+9.3), validating our specialized RL-based training approach. As shown in~\Cref{tab:refine}, while the powerful Wan2.5 achieves a higher performance (63.4), our trained model's substantial enhancement proves that our method effectively boosts open-source base models, offering a viable path for customized refinement solutions.

\noindent\textbf{Refining rounds.}
We analyze the impact of iterative refinement rounds in \Cref{tab:round}.
The results demonstrate that the refinement model's capability is critical. The base Qwen-Image-Edit model shows performance degradation with each round (Score 54.3 $\rightarrow$ 49.4), indicating an inability to process iterative corrections. In contrast, our trained model (Qwen-Image-Edit-2509*) maintains stable improvement (53.7 $\rightarrow$ 54.9), validating that our specialized training successfully addresses this error accumulation.
Furthermore, the powerful Wan2.5 model achieves continuous improvement (61.3 $\rightarrow$ 63.4), confirming that more capable models are essential for effective multi-round refinement. This underscores the necessity of our training, which successfully enhances the open-source model to reliably support the iterative process.

\section{Conclusion}
This work introduces the creative table visualization task that demands both aesthetic graphic reasoning and high-fidelity data mapping, addressing the critical challenge of generating faithful and aesthetic data visualizations. To address this, we propose \textbf{ShowTable}, a novel pipeline that synergizes MLLMs with diffusion models through iterative reflection and refinement, significantly improving visual-data mapping alignment. To support this task, we also present three automated data construction pipelines for different module training. Moreover, we propose a new comprehensive benchmark \textbf{TableVisBench} with 5 evaluation dimensions. Experiments demonstrate our approach's effectiveness in producing accurate and aesthetically coherent table visualizations, establishing a foundation for future research in multi-modal reasoning and visual synthesis.

\section*{Acknowledgment}
This work is supported by the National Nature Science Foundation of China (62425114, 62121002, U23B2028), and the Fundamental and Interdisciplinary Disciplines Breakthrough Plan of the Ministry of Education of China (JYB2025XDXM103). We acknowledge the support of Alibaba Group, the GPU cluster built by MCC Lab of Information Science and Technology Institution, USTC, and USTC super-computing center for providing computational resources for this project.

{
    \small
    \bibliographystyle{ieeenat_fullname}
    \bibliography{main}
}


\appendix
\clearpage

\newcommand\DoToC{%
    \startcontents
    \printcontents{}{1}{\hrulefill\vskip0pt}
    \vskip0pt \noindent\hrulefill
    }

\setcounter{page}{1}
\setcounter{table}{0}
\setcounter{figure}{0}
\setcounter{equation}{0}
\setcounter{footnote}{0}
\renewcommand{\thetable}{A\arabic{table}}
\renewcommand{\thefigure}{A\arabic{figure}}
\renewcommand{\theequation}{A\arabic{equation}}

\twocolumn[{\begin{center}
    \Large
    \textbf{\thetitle}\\
    \vspace{0.5em}Supplementary Material \\
    \vspace{1.0em}
\end{center}
}]

\noindent\textbf{Overview}
\noindent\DoToC

\section{More Related Works}
\subsection{Chart Generation with Agentic Tools}
With rapid development of current MLLMs~\cite{llava, qwen2.5vl, liu2025hybrid, zheng2025deepeyes}, the strong tool calling ability provides a promising way to complete multi-modal tasks. Current approaches for chart generation predominantly rely on LLMs coupled with agentic tools, generally falling into three main categories. 
The first category~\cite{tian2024chartgpt,rashid2022text2chart,zhang2024chartifytext} employs LLMs in an end-to-end manner to produce charts directly from text. These methods typically parse input to identify axes, map entities, and classify chart types, subsequently generating structured specifications for rendering engines like Vega-Lite. While automated, their expressiveness is strictly confined by the predefined grammatical rules of the underlying visualization language.
The second category~\cite{zhang2025thyme,zhao2025pyvision,zhao2025chartedit,zadeh2024text2chart31,han2023chartllama} focuses on generating executable plotting code (\eg, via Matplotlib/Python). Recent works have incorporated multi-agent frameworks and reflection mechanisms to improve the syntactic correctness of the generated code~\cite{ford2025does}. Although these methods offer precise control, they heavily depend on external rendering engines and typically lack the capability to handle complex, artistic visual designs beyond standard plots.
The third category~\cite{xiao2023let} adopts a retrieval-editing pipeline that selects visual templates from an image corpus and adapts them to new data. While this benefits from template reuse, it is limited by the diversity of the database and often faces challenges in accurately aligning new data with retrieved visual structures.

\begin{table}[!t]
\setlength\tabcolsep{3pt}
\centering
\caption{The comparison between our ShowTable pipeline and code-based methods. For our ShowTable pipeline, we use Gemini-2.5-pro for rewriting, Wan2.5-T2I-Preview for generation, GPT-5 for Reflection, and Wan2.5-I2I-Preview for refinement here. For code-based methods, we use Gemini-2.5-pro to write python code with the matplotlib library.}
\label{tab:supp_comp_with_code}
\resizebox{0.48\textwidth}{!}{
\begin{tabular}{l|ccccc|c}
\toprule
\textbf{Methods} & \textbf{DA} & \textbf{TR} & \textbf{RR} & \textbf{AA} & \textbf{AQ} & \textbf{Score} \\ \midrule
ShowTable Pipeline & 69.6 & 84.9 & 67.6 & 68.4 & 4.9 & 67.9 \\
Code (Gemini-2.5-pro) & 83.8 & 89.2 & 85.3 & 97.7 & 4.1 & 79.4 \\ \bottomrule
\end{tabular}%
}
\end{table}

\begin{figure}[!t]
    \centering
    \includegraphics[width=0.48\textwidth]{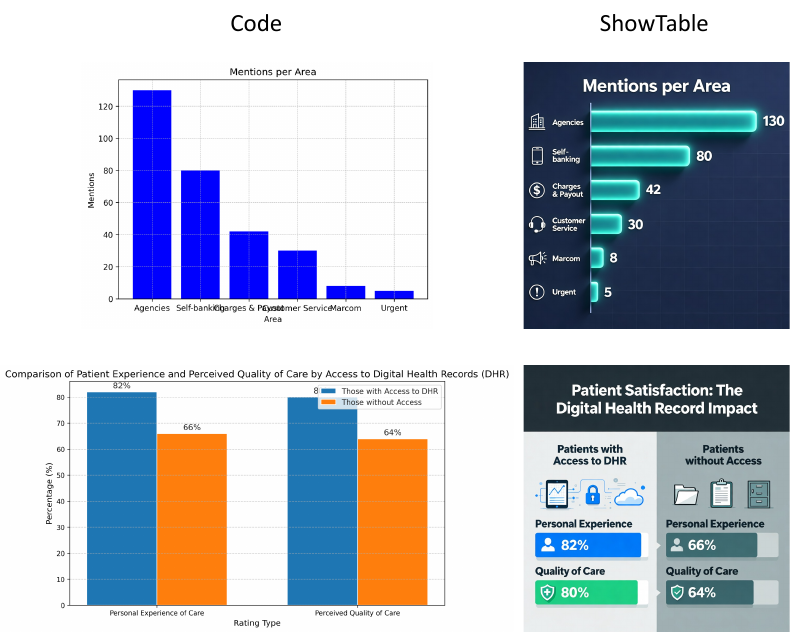}
    \caption{The qualitative comparison between code-based methods and our ShowTable pipeline. The code-based methods often prioritize ``correctness'' but may fail at ``presentation''}
    \label{fig:SM_comp_with_code}
\end{figure}

\noindent\textbf{Discussion.}  
While existing methods excel in structural automation and factual plotting, they fundamentally struggle with \textbf{creativity} and \textbf{aesthetics}. Code-based and template-based approaches are bound by rigid rendering logic, making it difficult to produce visually striking infographics suitable for professional poster design, slide generation, or data-driven storytelling. They prioritize ``correctness'' but often fail at ``presentation''.
To explicitly validate this, we conducted a quantitative comparison between a strong code-based baseline (e.g., using Gemini-2.5-Pro to generate executable plotting code) and our ShowTable pipeline, as shown in~\cref{tab:supp_comp_with_code} and ~\cref{fig:SM_comp_with_code}. As expected, code generation methods inherently achieve superior factual accuracy across structural metrics such as Data Accuracy and Relative Relationship. However, they fall significantly short in Aesthetic Quality (AQ: 4.1 vs. our 4.9) and frequently produce visual artifacts, such as overlapping text. This rigidity restricts their utility in design-centric, creative scenarios.

Importantly, our objective is not to replace traditional code-based rendering tools where absolute numerical precision is the sole priority. Instead, our work proposes the Creative Table Visualization task to explore the untapped potential of generative models and unified models in this domain. We frame our contribution as a technical advancement in multi-modal generation control: we aim to retain the unparalleled creative flexibility of generation models while actively mitigating their primary weakness, data fidelity, through our progressive self-correcting pipeline.
We argue that generative models offer a significantly higher ceiling for flexibility and aesthetic quality, capable of seamlessly integrating data into artistic compositions. Furthermore, equipping text-to-image models with rigorous data fidelity is a vital frontier for achieving more general capabilities in visual synthesis (including scientific reporting), as seen in the trajectory of recent models like Wan2.6~\cite{wan26} and Nano Banana~\cite{nano_banana} after the CVPR submission. By validating the feasibility of this approach, we aim to break the traditional boundaries of rule-based rendering and pave the way for more robust, unified, and creative visual synthesis systems.

\subsection{Reinforcement Learning for Image Generation}
Diffusion models have established themselves as the predominant framework for text-to-image (T2I) generation~\cite{flux, sd, podell2023sdxl, ddpm, qwenimage, jiang2025enhancing} and text-to-video (T2V) generation~\cite{jiang2023efficient, wei2024dreamvideo, wei2024dreamvideo2, wei2025dreamrelation}. The integration of reinforcement learning (RL) into this paradigm began with works utilizing policy gradient optimization to guide the denoising process~\cite{ddpo, rewarddance, imagereward, dpok}. The field subsequently expanded to include preference-based alignment methods, which achieve competitive performance without explicit reward modeling~\cite{dpo}. A significant recent development is the adoption of Group Relative Policy Optimization(GRPO)~\cite{deepseekr1, deepseekmath}, an efficient alternative that has inspired numerous adaptations for T2I generation. These include pioneering works~\cite{dancegrpo, flowgrpo} which unified diffusion and flow matching under an SDE-based formulation. This line of inquiry further explores the design of specialized reward models and data curation strategies to enhance the framework's capability for producing high-quality, preference-aligned visual outputs~\cite{mixgrpo, prefgrpo, branchgrpo}.

\section{More Method Details}
\subsection{Prompts in Pipeline}

\begin{figure*}[!t]
    \centering
    \includegraphics[width=0.9\textwidth]{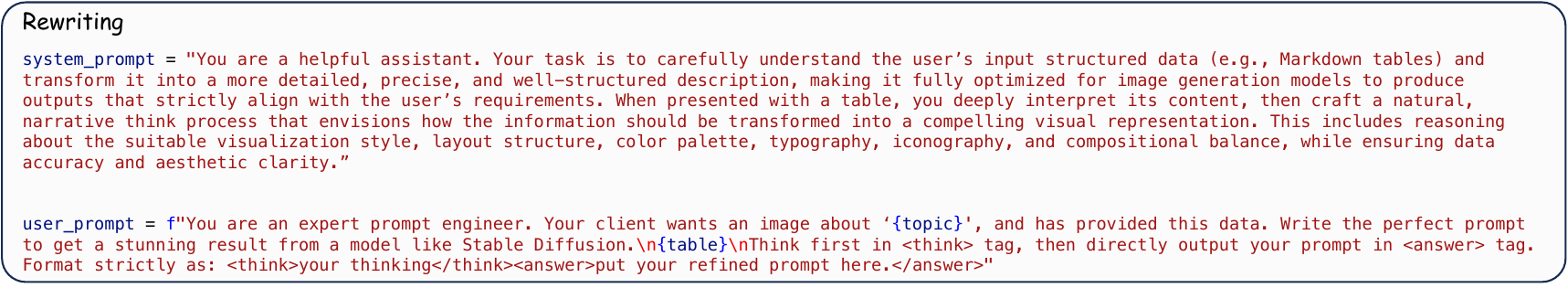}
    \caption{The system prompt and user prompt for the rewriting module. We use the same prompts for all models.}
    \label{fig:rewrting_prompt}
\end{figure*}

\begin{figure*}[!t]
    \centering
    \includegraphics[width=0.9\textwidth]{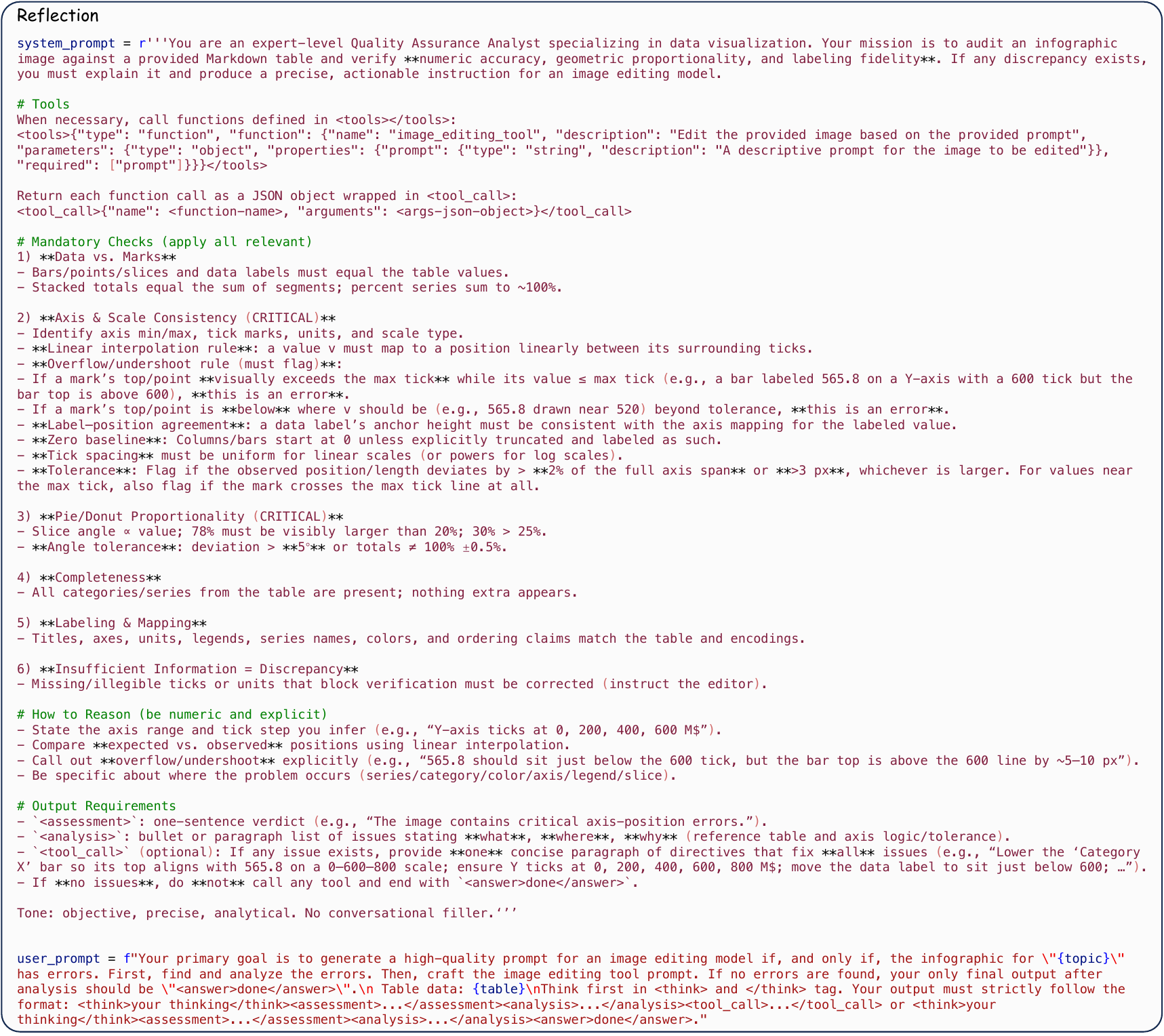}
    \caption{The system prompt and user prompt for the reflection module. We use the same prompts for all models.}
    \label{fig:reflection_prompt}
\end{figure*}

In our ShowTable pipeline, the MLLM acts as the central orchestrator, performing two key roles: rewriting and reflection. We detail the specific prompts used for these modules. The rewriting prompt, shown in~\Cref{fig:rewrting_prompt}, instructs the MLLM to reason over the input table and translate its dense data into a detailed descriptive prompt suitable for the diffusion executor. The reflection prompt, shown in \Cref{fig:reflection_prompt}, guides the MLLM to critically audit the generated image against the original table, identify inaccuracies, and formulate precise, actionable editing instructions for the refinement stage. To ensure fair comparison and consistency across experiments, we use the same system and user prompts for all models tested in these roles.

\subsection{Rewriting Training Details}
As discussed in~\Cref{sec:training_detail}, we fine-tune a specialized rewriting module based on Qwen3-8B to handle the critical task of reasoning and compositional planning. This module is trained on our 30K SFT data with both thinking and rewriting result, and we construct various kinds of instruction templates during training to ensure the diversity. For the implementation, we utilize the LLaMA-Factory~\cite{llamafactory} library. We train the model for 3 epochs with a total batch size of 256. The training employs a learning rate of 1e-5, combined with a cosine learning rate decay strategy to stabilize the training process.

Through targeted fine-tuning, the resulting rewrite model acquires the ability to infer appropriate data visualization and layout strategies. The thinking process enables the model to generate significantly improved prompts, thereby enhancing both data integrity and accuracy throughout the prompt rewriting and subsequent image generation pipeline.

\begin{figure}[!t]
    \centering
    \includegraphics[width=0.45\textwidth]{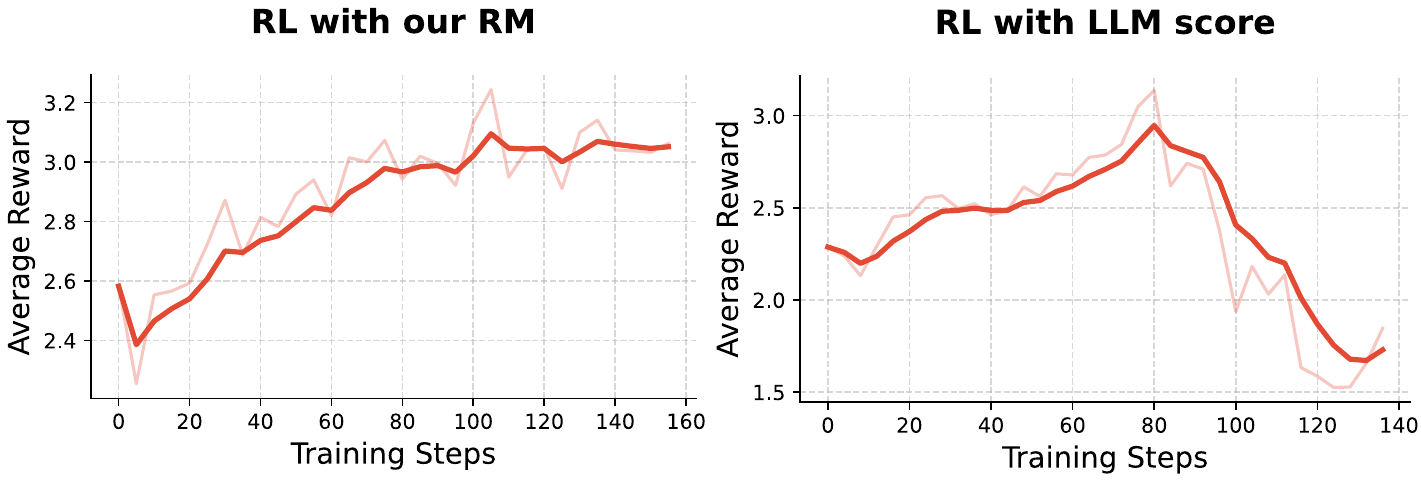}
    \caption{The reward comparison between our RM and direct LLM scoring, we achieve a stable reward increase.}
    \label{fig:reward_curve}
\end{figure}

\subsection{Refinement Training Details}
To empower the refinement model with the capability for precise, fine-grained edits on dense data infographics, we employ an on-policy reinforcement learning approach utilizing the Group Relative Policy Optimization (GRPO)~\cite{deepseekr1} algorithm. In this framework, rendering accuracy serves as the critical reward signal.

\noindent\textbf{Reward model.}
Evaluating the rendering accuracy of infographics is complex, requiring the integration of multiple dimensions—such as textual correctness, data-to-visual alignment, and spatial layout. Our preliminary experiments revealed that utilizing state-of-the-art pretrained Vision-Language Models (VLMs) directly for point-wise reward assessment is suboptimal. As shown in~\Cref{fig:reward_curve}, the inconsistency in VLM scoring often leads to training instability or collapse. Furthermore, the high inference latency of VLMs significantly hampers the efficiency of the on-policy GRPO loop.

To address this, we develop a specialized, efficient Reward Model (RM). We construct a pairwise dataset $D$ consisting of positive and negative graphic samples generated from the same prompt, as detailed in~\Cref{sec:train_data}. We fine-tune a Qwen2.5-VL-3B~\cite{qwen2.5vl} model, denoted as $f_\theta$, to serve as the quality assessor. For a given prompt $p$, with $x_w$ denoting the preferred (positive) sample and $x_l$ the dispreferred (negative) sample, the model is optimized using the Bradley-Terry (BT) loss~\cite{bradley1952bradterry}:
\begin{equation}
    \mathcal{L}_{BT} = -\mathbb{E}_{(p, x_w, x_l) \sim D} \left[ \log \sigma \left( f_\theta(x_w, p) - f_\theta(x_l, p) \right) \right],
\end{equation}
where $\sigma(\cdot)$ represents the Sigmoid function. To stabilize the output, the reward score is computed by extracting and averaging the probabilities corresponding to the tokens for digits 0–9 from the output logits. The final reward signal used in RL is a weighted combination: $R = 0.8 \cdot f_\theta(x, p) + 0.2 \cdot \text{ImageReward}(x, p)$~\cite{imagereward}.

\noindent\textbf{Policy optimization.}
The refinement policy is updated using the GRPO algorithm. The objective function maximizes the expected reward while constraining policy divergence via a clipped surrogate objective:
\begin{equation}
\begin{split}
    J_{\text{GRPO}}(\theta) ={}& \frac{1}{G} \sum_{i=1}^{G} \mathbb{E}_{y \sim D} \Biggl[ \min\left( \frac{\pi_\theta(o_i|y)}{\pi_{\theta_{\text{old}}}(o_i|y)} A_i, \right. \\
    & \quad \left. \text{clip}\left( \frac{\pi_\theta(o_i|y)}{\pi_{\theta_{\text{old}}}(o_i|y)}, 1-\varepsilon, 1+\varepsilon \right) A_i \right) \\
    & \quad - \beta D_{KL}\left( \pi_\theta \parallel \pi_{\text{ref}} \right) \Biggr],
\end{split}
\label{eq:grpo}
\end{equation}
where $\varepsilon$ and $\beta$ are hyperparameters, and $G$ is the group size. The advantage $A_i$ is computed by normalizing the rewards $\{r_1, r_2, \cdots, r_{G}\}$ within each group:
\begin{equation}
    A_i = \frac{r_i - \text{mean}(\{r_1, \cdots, r_{G}\})}{\text{std}(\{r_1, \cdots, r_{G}\}) + \epsilon}.
\end{equation}

\noindent\textbf{Implementation.}
Following the Flow-GRPO framework~\cite{flowgrpo}, we train our refinement model using a distilled 8-step version of Qwen-Image-Edit-2509~\cite{qwenimagelightning} to accelerate training efficiency. The model is trained on our constructed 5K refinement dataset for 1 epoch using 32 GPUs. We set the image resolution to $1024 \times 1024$ and perform 16 rollouts per prompt. We utilize 8 sampling steps for both the inference rollout and the training backward pass. As illustrated in~\Cref{fig:reward_curve}, compared to the unstable baseline using raw LLM scores, our approach with the trained Reward Model achieves stable and consistent performance gains.

\begin{figure*}[!t]
    \centering
    \includegraphics[width=0.95\textwidth]{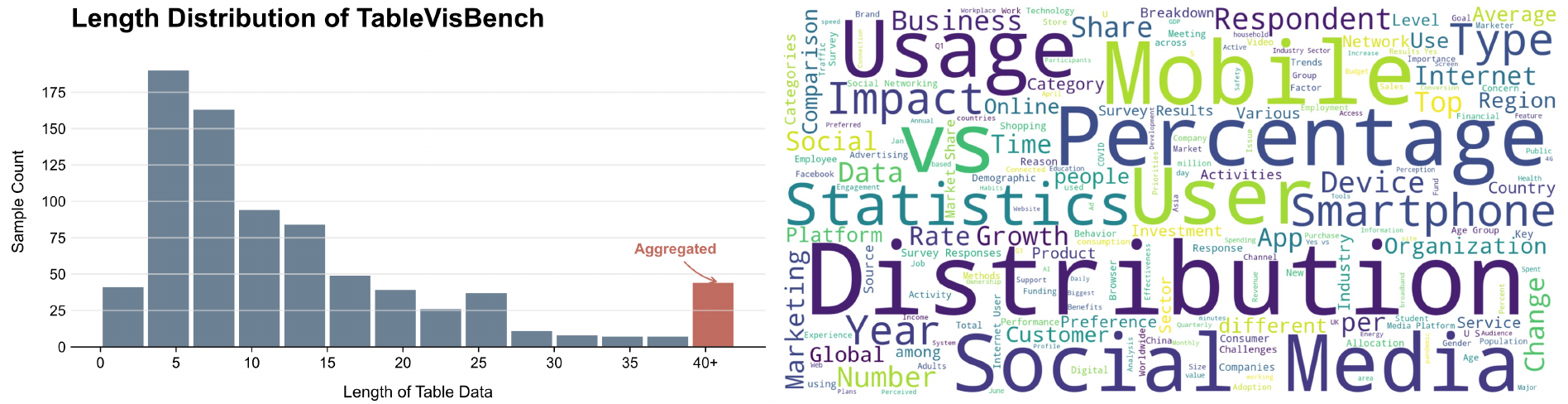}
    \caption{The Statistical information about our proposed TableVisBench. (a) (Left): the data length distribution of TableVisBench. (b) (Right): The word cloud of the topic in TableVisBench.}
    \label{fig:bench}
\end{figure*}

\section{More Dataset and Benchmark Details}
\subsection{Training Data Format}
As illustrated in~\Cref{fig:data_construct}, our data construction pipeline transforms raw collected images into specialized datasets tailored for the rewriting, refinement, and reward modules. Below, we detail the specific input and output formats for each training stage.

\noindent\textbf{Rewriting Training Data (SFT).}
The goal of the rewriting module is to convert a raw markdown table into a comprehensive visual plan. To support this, we construct Supervised Fine-Tuning (SFT) data that teaches the model to first reason about the data structure (Rationale) and then describe the visual elements (Description).
\begin{itemize}
    \item \textbf{Input:} The raw table data in markdown format, annotated from the collected image pool.
    \item \textbf{Output:} A composite text sequence consisting of a \textit{Chain-of-Thought Rationale} followed by a \textit{Detailed Image Description}.
    \item \textbf{Construction:} As shown in the green section of \Cref{fig:data_construct}, we first prompt Gemini-2.5-pro to generate a descriptive caption (``Table-based description''). Then, we feed both the table and the description into Gemini-2.5-pro again to reverse-engineer the reasoning process (``Rationale to description''), forming a complete training sample: $\text{Table} \rightarrow \text{Rationale} + \text{Description}$.
\end{itemize}

\noindent\textbf{Refinement Training Data (RL).}
For the reinforcement learning stage, the training data consists of challenging scenarios where an initial generation with correction instructions, and no ground-truth image is needed. This data is formatted as prompt-response pairs for the policy model.
\begin{itemize}
    \item \textbf{Input:} A pair consisting of an \textit{Initial Generated Image} (containing errors) and a precise \textit{Refinement Instruction}.
    \item \textbf{Output:} No refined image is needed.
    \item \textbf{Construction:} As shown in the blue section of \Cref{fig:data_construct}, we generate initial images from our descriptions and use Gemini-2.5-pro to compare them against the table, producing a ``Refine command.'' To ensure the data is valid for training, we perform a ``Rollout'' check: we discard samples where the base model either fails to improve the image over multiple attempts (too hard) or solves it trivially (too easy), retaining only those suitable for learning stable policy gradients.
\end{itemize}

\noindent\textbf{Reward Training Data (Preference Pairs).}
To train the reward model $f_\theta$ as a reliable quality assessor, we construct a dataset of preference image pairs, focusing on the data fidelity.
\begin{itemize}
    \item \textbf{Input:} A text condition (the table) and two candidate images ($x_w, x_l$).
    \item \textbf{Output:} A binary label indicating which image is the ``Winner'' ($x_w$) and which is the ``Loser'' ($x_l$).
    \item \textbf{Construction:} As shown in the purple section of \Cref{fig:data_construct}, we source candidates from three comparisons: (1) Refined vs. Initial images, (2) Strong (Wan2.5) vs. Weak (Qwen) model outputs, and (3) Ground-truth vs. Generated images. MLLMs (GPT-5 and Gemini-2.5-pro) act as judges to vote on the pair, establishing a high-confidence ``Pos-Neg Pair'' dataset for training the reward model to discriminate fine-grained visual differences.
\end{itemize}

\begin{figure*}[!t]
    \centering
    \includegraphics[width=0.9\textwidth]{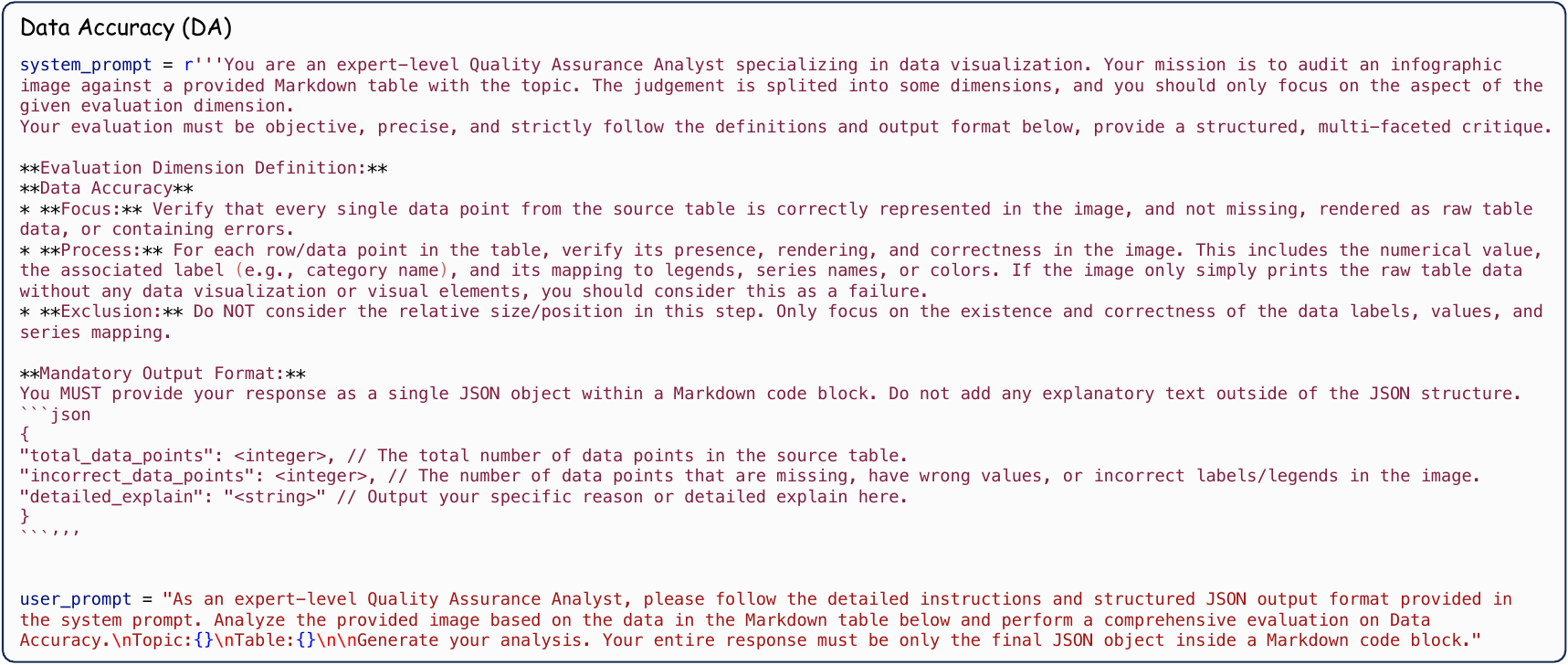}
    \caption{The system prompt and user prompt of the Data Accuracy dimension.}
    \label{fig:DA_prompt}
\end{figure*}

\begin{figure*}[!t]
    \centering
    \includegraphics[width=0.9\textwidth]{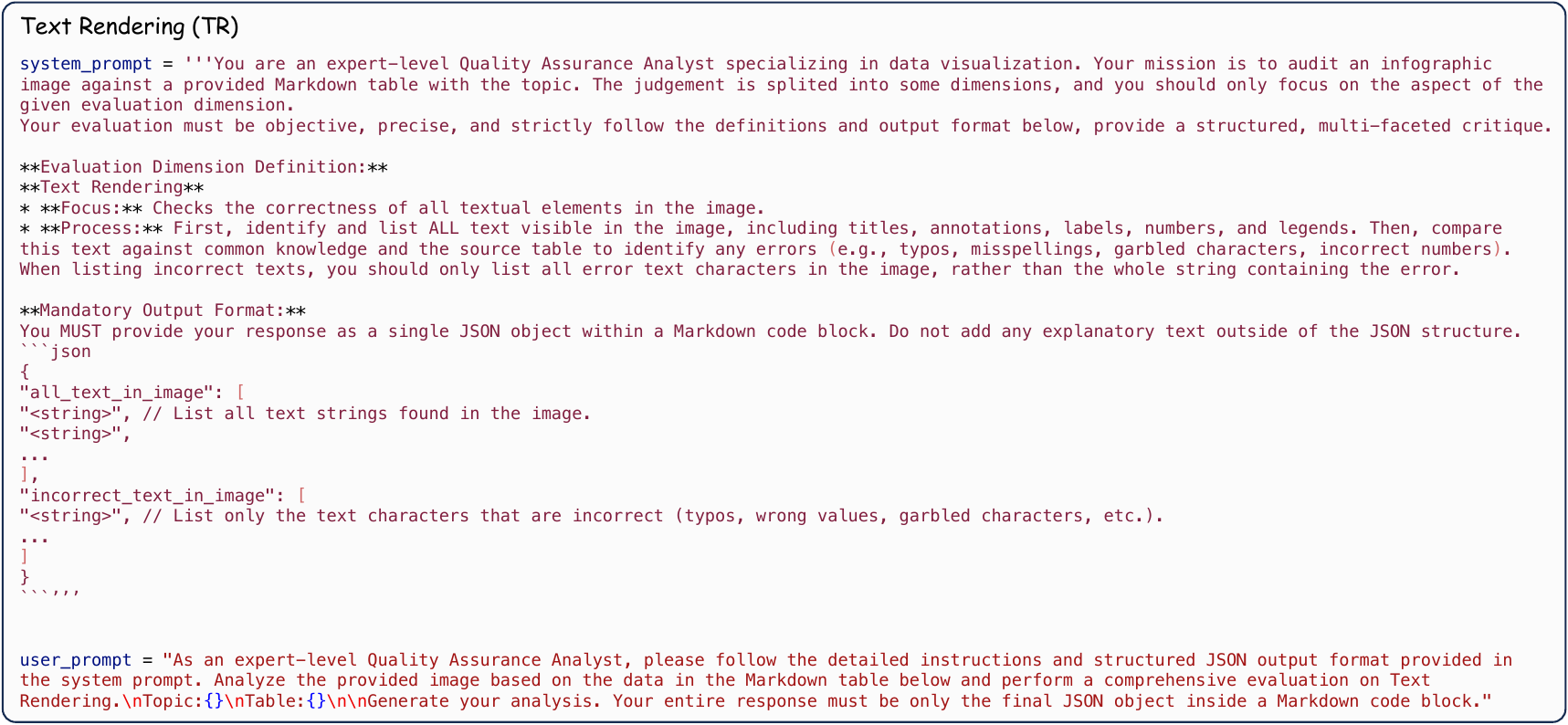}
    \caption{The system prompt and user prompt of the Text Rendering dimension.}
    \label{fig:TR_prompt}
\end{figure*}

\begin{figure*}[!t]
    \centering
    \includegraphics[width=0.9\textwidth]{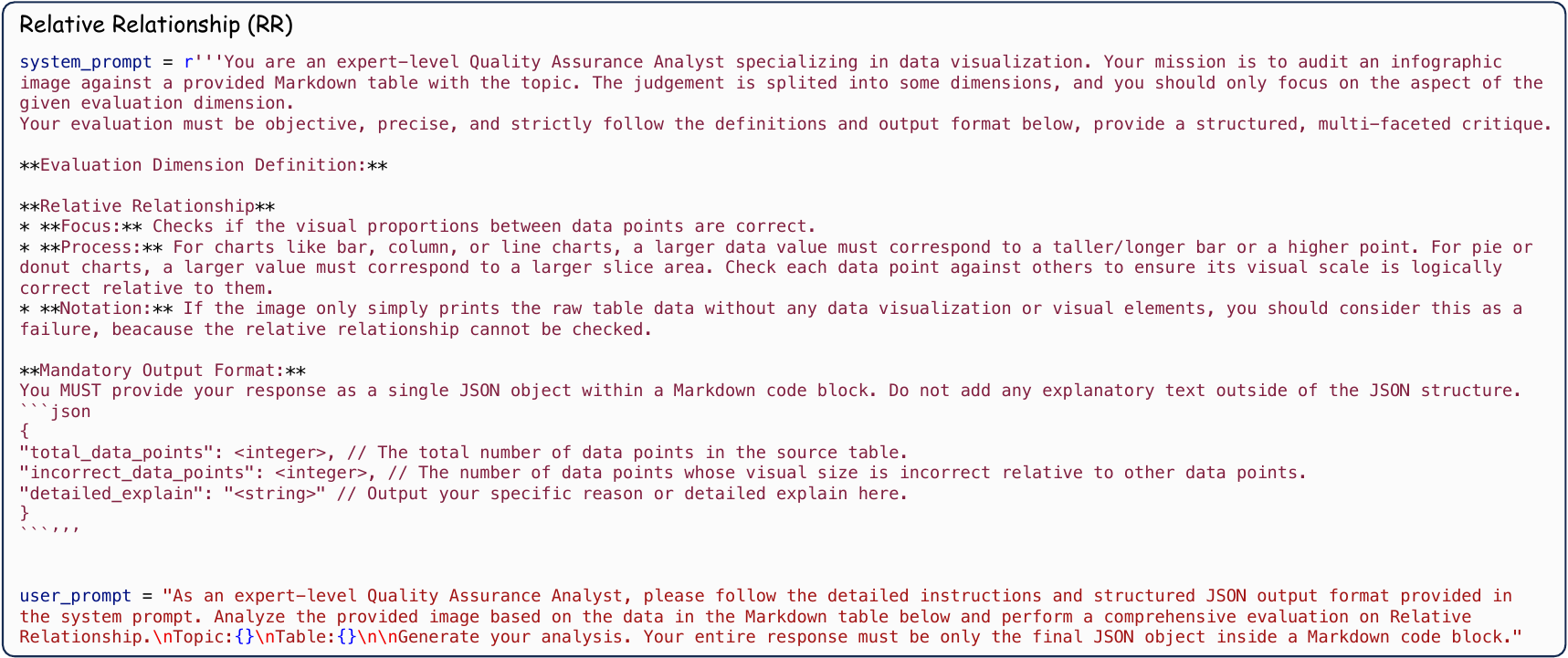}
    \caption{The system prompt and user prompt of the Relative Relationship dimension.}
    \label{fig:RR_prompt}
\end{figure*}

\begin{figure*}[!t]
    \centering
    \includegraphics[width=0.9\textwidth]{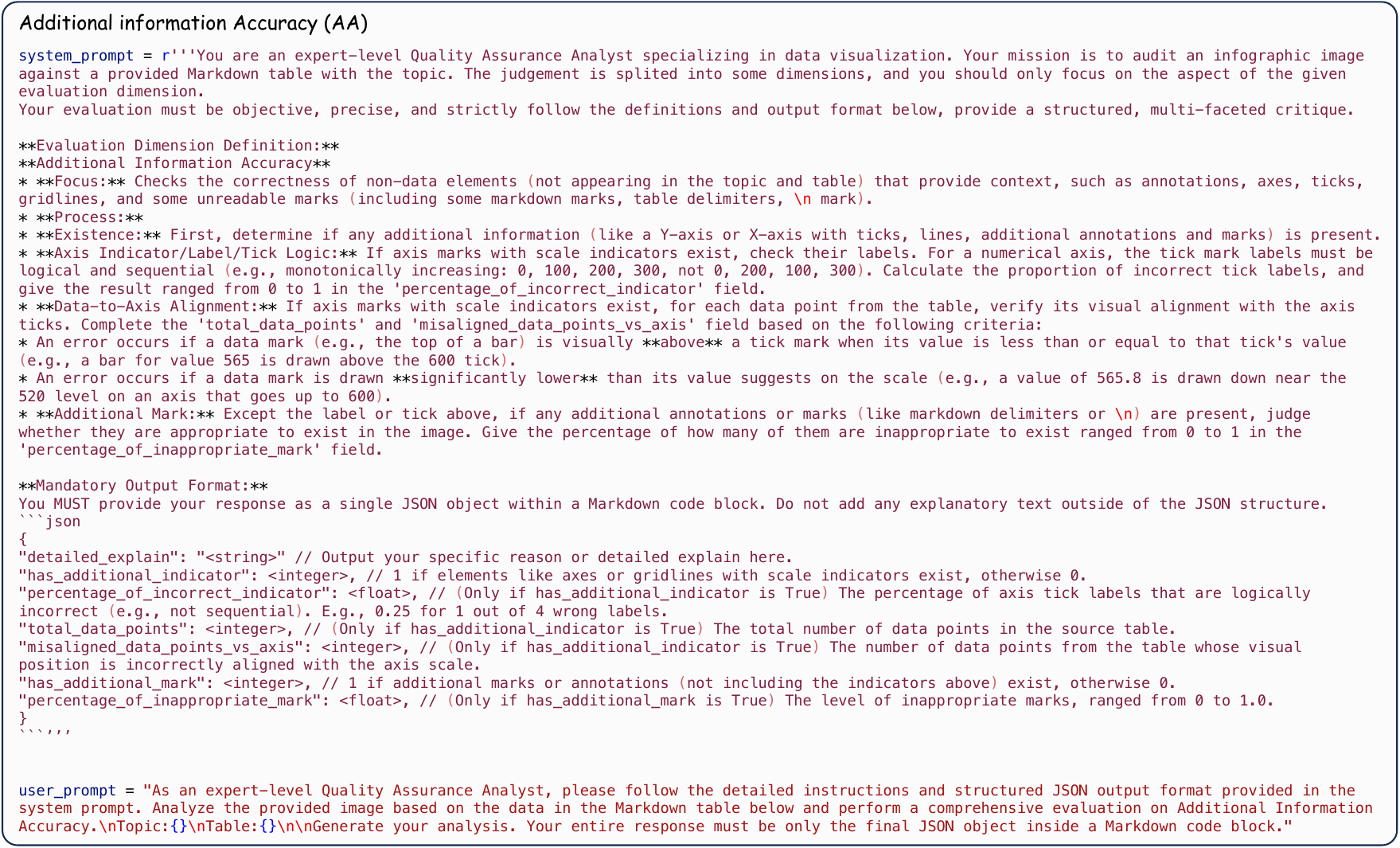}
    \caption{The system prompt and user prompt of the Additional information Accuracy dimension.}
    \label{fig:AA_prompt}
\end{figure*}

\subsection{Benchmark Statistics}
To demonstrate the diversity and complexity of our proposed TableVisBench, we present detailed statistical characteristics in \Cref{fig:bench}.

\noindent\textbf{Data Length Distribution.}
\Cref{fig:bench} (a) illustrates the distribution of table lengths, defined as the number of key data points per instance. The benchmark covers a broad spectrum of information density, ranging from concise tables (fewer than 5 rows) to highly complex ones. The distribution shows a natural concentration between 5 to 15 data points, reflecting common real-world infographic scenarios. Notably, we also include some ``long-tail'' instances, with the final bin aggregating tables containing over 40 data points. This design ensures that the benchmark rigorously evaluates models not only on standard visualizations but also on their stability and layout planning capabilities when handling high-density data.

\noindent\textbf{Topic Diversity.}
\Cref{fig:bench} (b) presents a word cloud visualization derived from the topics of the collected tables. The dataset encompasses a wide array of domains, with prominent keywords including ``Social Media,'' ``Distribution,'' ``Percentage,'' ``User,'' ``Market,'' and ``Mobile.'' This semantic diversity confirms that TableVisBench covers various distinct fields—such as business reports, sociological statistics, and technology usage—thereby assessing the model's generalization ability across different contexts and terminologies.

\subsection{Benchmark Evaluation Details}
To ensure a rigorous and reproducible evaluation, we design a deterministic scoring mechanism for our TableVisBench. Instead of asking the MLLM to directly output a subjective score (e.g., 1-10), we employ the MLLM as a \textit{Quality Assurance Analyst} to identify and count specific errors based on strict definitions. The final scores are calculated deterministically from these counts. Below are the detailed calculation protocols for the four accuracy-based dimensions.

\noindent\textbf{Data Accuracy (DA).}
This dimension measures the completeness and correctness of the data points rendered. The MLLM identifies the total number of data points in the source table ($N_{total}$) and counts the number of incorrect data points ($N_{error}$) in the image (including missing values, wrong numbers, or incorrect legend mappings). The specific prompt is shown in~\Cref{fig:DA_prompt}. The score is calculated as:
\begin{equation}
    S_{DA} = \frac{N_{total} - N_{error}}{N_{total}}
\end{equation}

\noindent\textbf{Text Rendering (TR).}
This dimension evaluates the character-level correctness of textual elements. The MLLM extracts all visible text strings from the image and identifies specific substrings or characters that contain errors (\eg, typos, garbled text). The specific prompt is shown in~\Cref{fig:TR_prompt}. Let $L_{total}$ be the total character length of all text in the image, and $L_{error}$ be the total character length of the identified incorrect text. The score is defined as:
\begin{equation}
    S_{TR} = \frac{L_{total} - L_{error}}{L_{total}}
\end{equation}
If no text is present ($L_{total}=0$), the score is set to 0.

\noindent\textbf{Relative Relationship (RR).}
This dimension assesses the visual proportionality of the infographic (e.g., whether bar heights or pie slice angles correspond to the data values). Similar to DA, the MLLM counts the total data points ($N_{total}$) and identifies the number of points ($N_{error}$) that violate visual logic relative to other elements. The specific prompt is shown in~\Cref{fig:RR_prompt}. The score is calculated as:
\begin{equation}
    S_{RR} = \frac{N_{total} - N_{error}}{N_{total}}
\end{equation}

\begin{table*}[!t]
\centering
\caption{Performance of Wan2.5-Preview on our TableVisBench.}
\label{tab:main_wan2.5}
\resizebox{0.9\textwidth}{!}{
\begin{tabular}{lccc|ccccc|c}
\toprule
\textbf{Rewriting}&\textbf{Generation}&\textbf{Reflection}&\textbf{Refinement}& \textbf{DA} & \textbf{TR}&\textbf{RR}&\textbf{AA}&\textbf{AQ}&\textbf{Score} \\
\midrule
\multicolumn{4}{l|}{\textit{Reference Image}}&\textit{97.7}&\textit{99.5}&\textit{86.4}&\textit{96.6}&\textit{4.2}&\textit{84.4}\\
\midrule
-&Wan2.5-T2I-Preview&-&-&62.8&82.5&62.8&56.5&4.8&62.5\\
Qwen3-8B*&Wan2.5-T2I-Preview&-&-&73.3&92.9&70.5&79.7&4.4&72.1\\
\midrule
Qwen3-8B*&Wan2.5-T2I-Preview&Gemini-2.5-pro&Qwen-Image-Edit-2509&53.3&88.3&53.5&59.5&4.3&59.5\\
Qwen3-8B*&Wan2.5-T2I-Preview&Gemini-2.5-pro&Qwen-Image-Edit-2509*&68.9&91.7&64.9&71.2&4.4&68.1\\
Qwen3-8B*&Wan2.5-T2I-Preview&Gemini-2.5-pro&Wan2.5-I2I-Preview&76.9&91.1&74.3&74.3&4.4&72.1\\
\bottomrule
\end{tabular}%
}
\end{table*}

\noindent\textbf{Additional information Accuracy (AA).}
This dimension evaluates contextual elements such as axes, ticks, and extra annotations. The score is an average of up to three sub-metrics, depending on the elements present in the image:
\begin{enumerate}
    \item \textbf{Label Logic ($S_{lbl}$):} If axis indicators exist, the model estimates the percentage of incorrect tick labels ($P_{err}$). Then $S_{lbl} = 1 - P_{err}$.
    \item \textbf{Axis Alignment ($S_{align}$):} If axes exist, the model checks the visual alignment of data points against axis ticks. Let $N_{mis}$ be the number of misaligned points. Then $S_{align} = \frac{N_{total} - N_{mis}}{N_{total}}$.
    \item \textbf{Artifact Appropriateness ($S_{mark}$):} If other marks (e.g., Markdown delimiters, random symbols) exist, the model estimates the percentage of inappropriate marks ($P_{inapp}$). Then $S_{mark} = 1 - P_{inapp}$.
\end{enumerate}
The specific prompt is shown in~\Cref{fig:AA_prompt}. The final $S_{AA}$ is the arithmetic mean of the valid sub-metrics. If no additional information is present, this dimension is excluded from the calculation.

\noindent\textbf{Aesthetic Quality (AQ).}
Unlike the strictly factual dimensions above, this dimension evaluates the overall visual appeal, including layout harmony, color palette suitability, and typographic quality. Since aesthetic judgment relies on subjective perception rather than rule-based error counting, we employ a dedicated pre-trained aesthetic scoring model to quantitatively assess this dimension. The model provides a scalar score $S_{AQ}$ ranging from 0 to 10, reflecting the artistic quality of the infographic independent of its data fidelity.

\section{More Experiments}
\subsection{Results of Wan2.5-Preview}
In the main paper, we primarily focused on open-source models to ensure reproducibility. Here, we extend our evaluation to the recently released Wan2.5-Preview series~\cite{wan2.5} to assess the ShowTable pipeline's performance with state-of-the-art (SOTA) generation capabilities. The results are presented in \Cref{tab:main_wan2.5}.

\noindent\textbf{Impact of rewriting.}
Consistent with our findings on open-source models, the rewriting module provides a substantial performance boost for Wan2.5-T2I-Preview. By converting the raw table into a reasoned visual plan (RW), the overall score increases significantly from 62.5 to 72.1 (+9.6). This improvement is particularly evident in Data Accuracy (62.8 $\rightarrow$ 73.3) and Text Rendering (82.5 $\rightarrow$ 92.9), confirming that our rewriting strategy is model-agnostic and effective even for top-tier generation models.

\noindent\textbf{Analysis of refinement.}
The refinement stage reveals the critical importance of the editing model's underlying capacity.
When applying the open-source \textit{Qwen-Image-Edit-2509} to refine images generated by the powerful \textit{Wan2.5-T2I-Preview}, we observe a performance degradation (Score 72.1 $\rightarrow$ 59.5) with the base editor. This is attributed to the significant capacity gap between the strong generator and the relatively weaker editor. Essentially, the editor struggles to maintain the high-fidelity details produced by the SOTA generator.
However, our proposed RL training demonstrates clear effectiveness in this challenging scenario. Our trained model (\textit{Qwen-Image-Edit-2509*}) significantly recovers performance compared to the base editor, raising the score from 59.5 to 68.1 (+8.6). 
Finally, when utilizing \textit{Wan2.5-I2I-Preview} as the executor, matching the generator's capacity, the pipeline achieves the highest performance in key structural metrics, with Data Accuracy improving further to 76.9 and Relative Relationship to 74.3. This underscores that while our training effectively boosts open-source editors, the ceiling of the refinement stage is ultimately determined by the base model's capability.

\subsection{More Visualization Results}

\begin{figure*}[!t]
    \centering
    \includegraphics[width=0.95\textwidth]{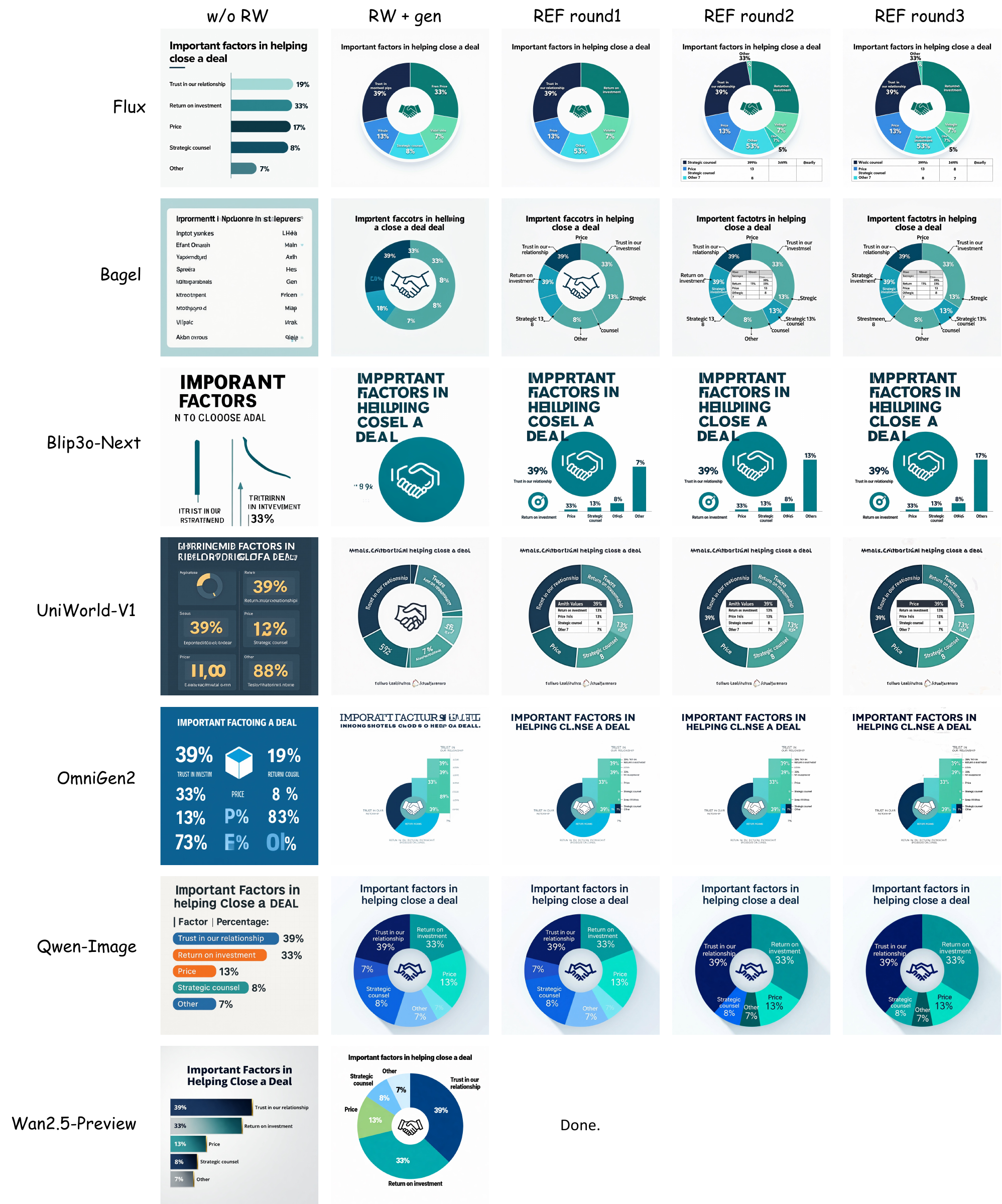}
    \caption{The qualitative comparison of different generation baselines of the same case on our proposed creative table visualization task. The first column presents the results without rewriting. The results via our ShowTable pipeline are shown from the second column.}
    \label{fig:SM_methods_case}
\end{figure*}

\noindent\textbf{Qualitative comparisons.}
To further qualitatively demonstrate the universality and effectiveness of our approach, we present a comprehensive case comparison across different base models in~\Cref{fig:SM_methods_case}.
As evident in the first column of \Cref{fig:SM_methods_case}, without the rewriting module, most base models fail to fundamentally grasp the visualization task. General unified models (e.g., Blip3o-Next, UniWorld-V1, OmniGen2) often suffer from severe hallucinations, producing chaotic layouts that lack any statistical meaning. Strong text-rendering models like Qwen-Image and Flux also fail to render the structure logits among data points. 
The second column (RW + gen) highlights the critical role of our rewriting module. By translating the data into a visual plan, all models successfully transition from unstructured chaos to a coherent infographic layout (specifically, a donut chart structure in this example), though most of them still suffer from heavy logical errors. Note that Wan2.5-Preview successfully generates a promising result, leading to no need of further refinement.
The subsequent columns (REF round 1-3) demonstrate the power of the progressive self-correction loop. Visual inaccuracies, such as incorrect data segmentation, garbled text, and layout misalignments, are iteratively repaired. For example, in the \textit{Bagel} and \textit{UniWorld-V1} rows, the text legibility and data mapping precision improve noticeably with each round. Additionally, the \textit{Wan2.5-Preview} case (bottom row) showcases the efficiency of our pipeline; due to its high initial quality, the reflection module triggers the early-stopping mechanism (``Done''), avoiding unnecessary computation.

\begin{figure*}[!t]
    \centering
    \includegraphics[width=0.95\textwidth]{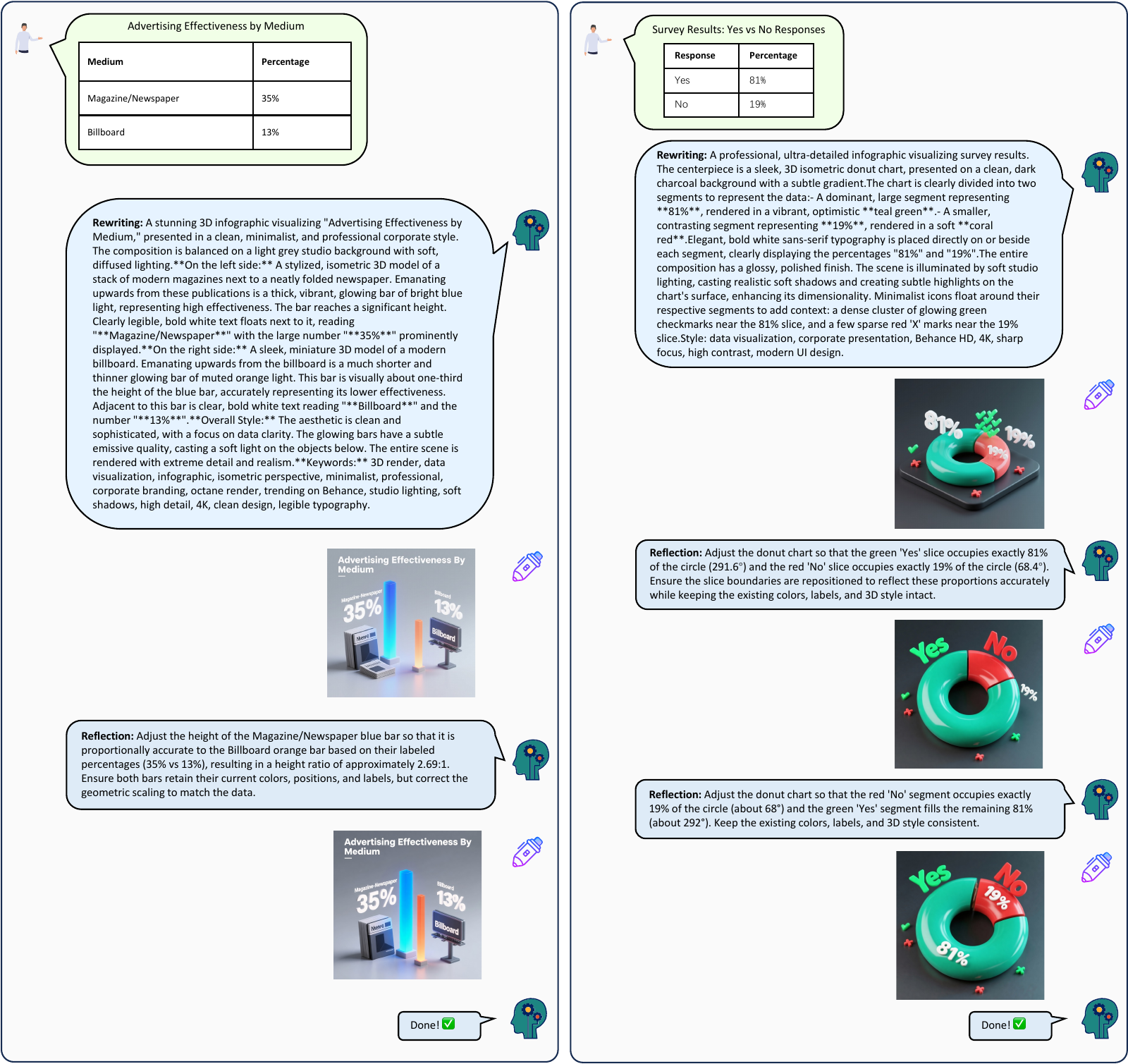}
    \caption{Some detailed pipeline visualizations. We use Qwen-Image here for base generation module with our pipeline.}
    \label{fig:SM_qwen_case1}
\end{figure*}

\begin{figure*}[!t]
    \centering
    \includegraphics[width=0.95\textwidth]{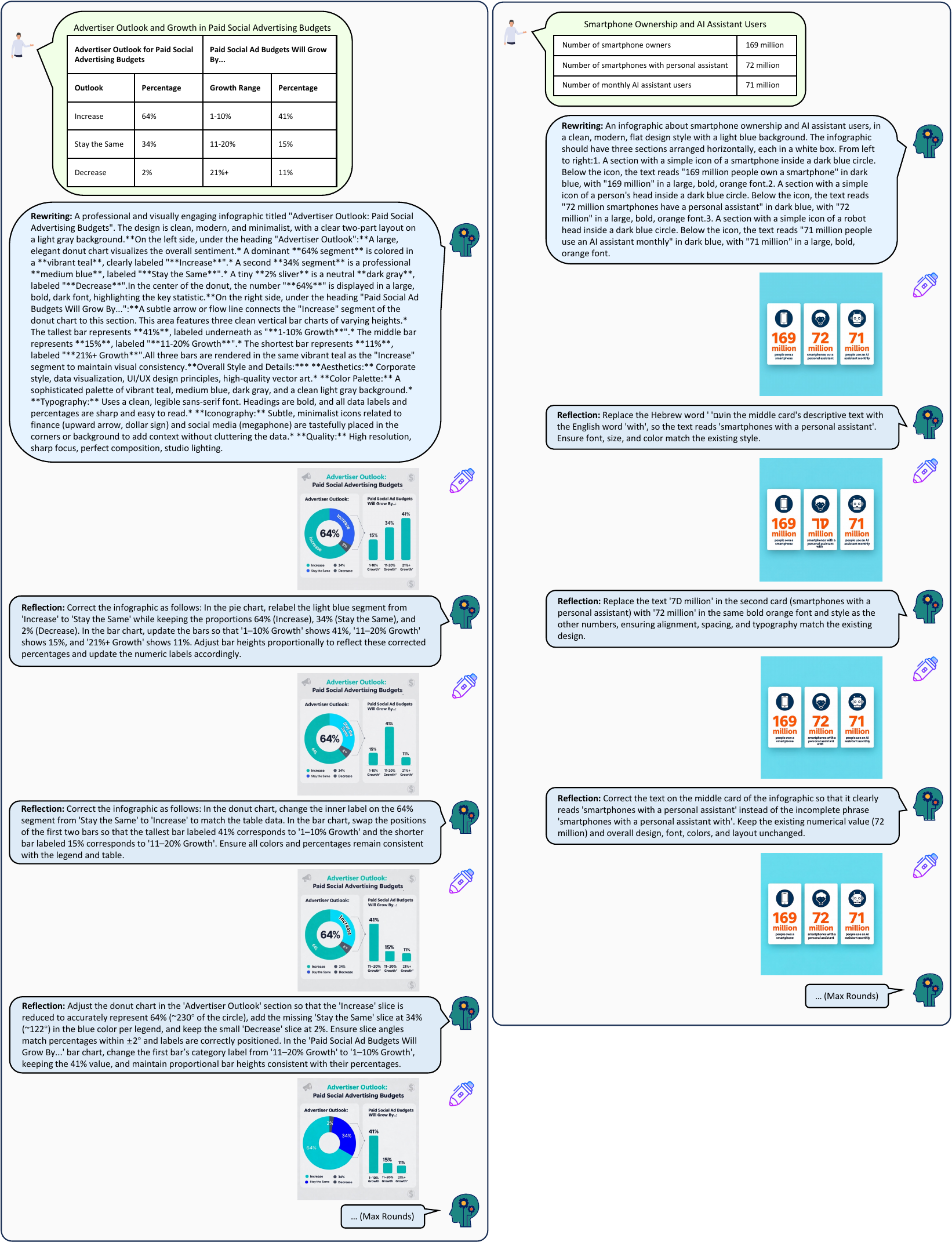}
    \caption{Some detailed pipeline visualizations. We use Qwen-Image here for base generation module with our pipeline.}
    \label{fig:SM_qwen_case2}
\end{figure*}

\begin{figure*}[!t]
    \centering
    \includegraphics[width=0.95\textwidth]{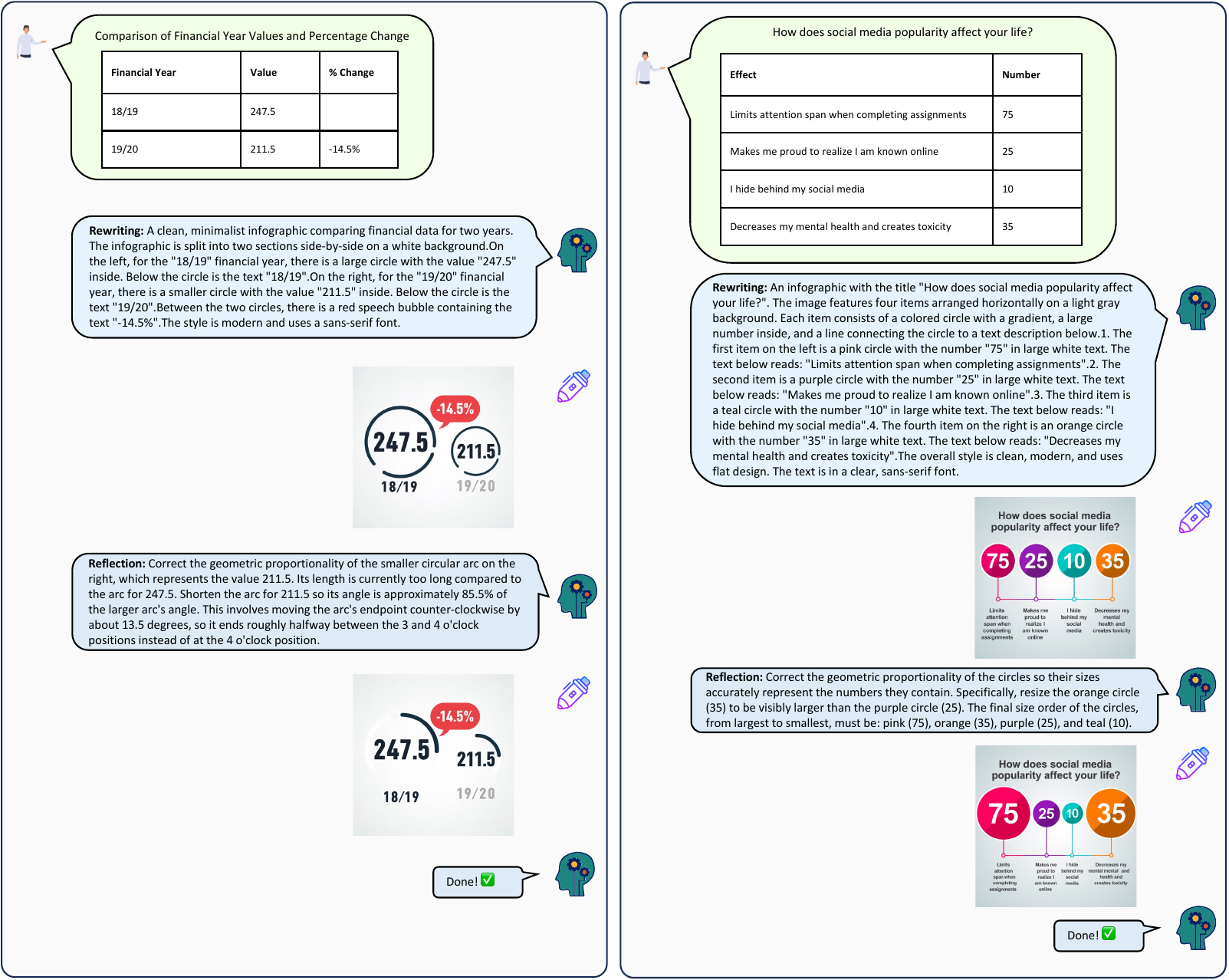}
    \caption{Some detailed pipeline visualizations. We use Qwen-Image here for base generation module with our pipeline.}
    \label{fig:SM_qwen_case3}
\end{figure*}

\begin{figure*}[!t]
    \centering
    \includegraphics[width=0.9\textwidth]{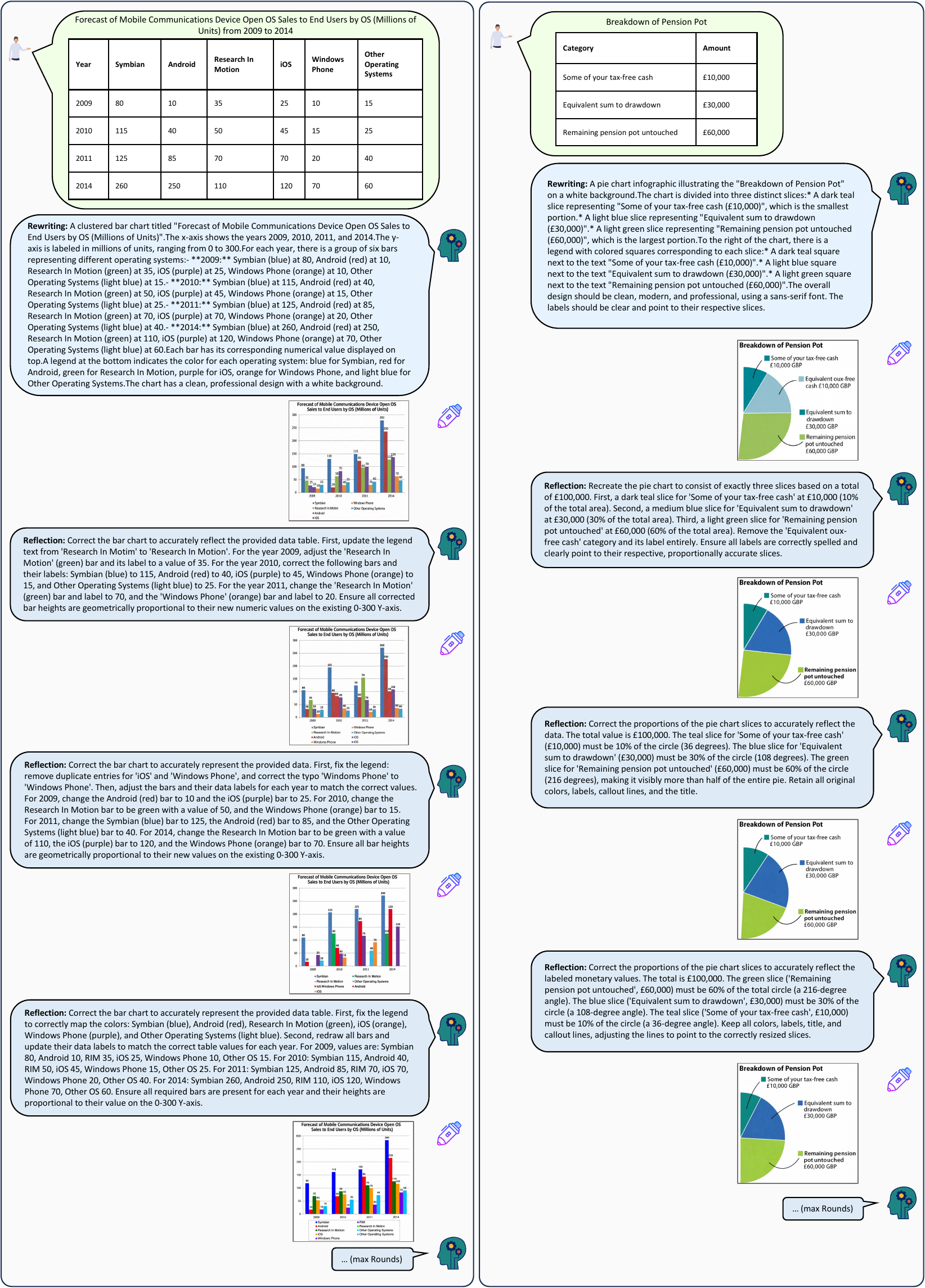}
    \caption{Some detailed pipeline visualizations with more complex and difficult cases. We use Wan2.5-Preview here for  generation module and refinement module with our pipeline.}
    \label{fig:SM_wan_case1}
\end{figure*}

\begin{figure*}[!t]
    \centering
    \includegraphics[width=0.95\textwidth]{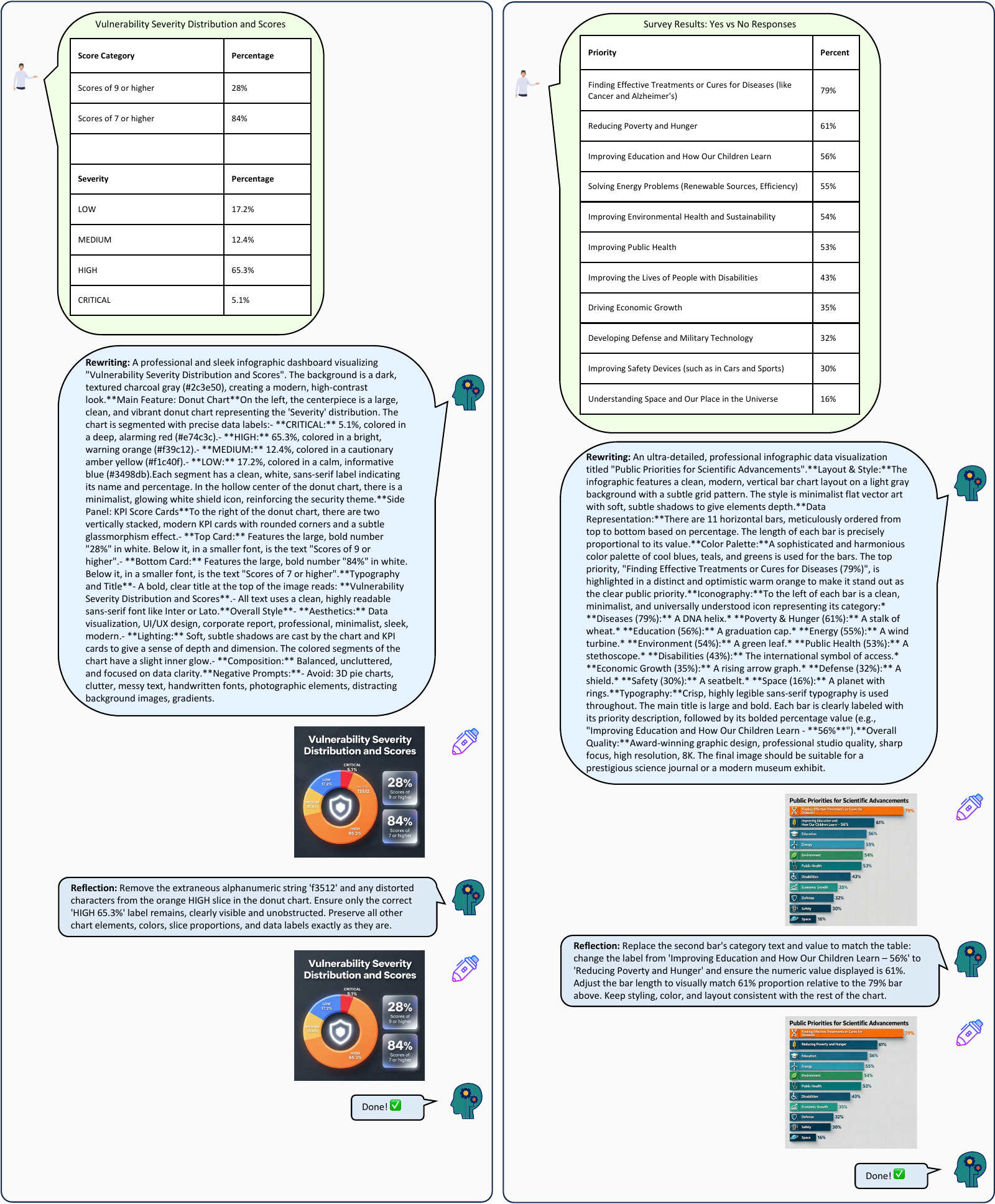}
    \caption{Some detailed pipeline visualizations with more complex and difficult cases. We use Wan2.5-Preview here for  generation module and refinement module with our pipeline.}
    \label{fig:SM_wan_case2}
\end{figure*}

\noindent\textbf{Detailed pipeline visualization.}
To provide a transparent view of the internal mechanisms of our ShowTable pipeline, we present a series of detailed case visualizations in the following figures. These figures explicitly display the step-by-step intermediate outputs, including the source table, the MLLM-generated \textbf{rewriting prompt}, the initial generation, and the iterative \textbf{reflection instructions} that guide the refinement.
To demonstrate the pipeline's versatility across different base models, the first three figures (\Cref{fig:SM_qwen_case1}, \Cref{fig:SM_qwen_case2}, \Cref{fig:SM_qwen_case3}) utilize \textbf{Qwen-Image} as the generation module, while the subsequent two figures (\Cref{fig:SM_wan_case1}, \Cref{fig:SM_wan_case2}) employ \textbf{Wan2.5-Preview}. As shown in the examples, the visualization highlights the orchestrator's ability to formulate precise geometric corrections (For example, ``adjust the height of the bar to... resulting in a ratio of approximately 2.69:1'' or ``ensure the green 'Yes' slice occupies exactly 81\% (291.6$^\circ$)''), effectively guiding the executor to achieve high-fidelity data alignment.

\begin{table}[!t]
\setlength\tabcolsep{3pt}
\centering
\caption{User study results showing human preference rates (\%) across five evaluation dimensions. Our full pipeline achieves the highest human preference in almost all dimensions, particularly dominating in Data Accuracy (DA).}
\label{tab:supp_user_study}
\resizebox{0.48\textwidth}{!}{
\begin{tabular}{lccccc}
\toprule
\textbf{Method} & \textbf{DA} & \textbf{TR} & \textbf{RR} & \textbf{AA} & \textbf{AQ} \\ \midrule
Qwen-Image & 2.25 & \textbf{38.50} & 12.00 & 2.50 & 7.75 \\
RW + Qwen-Image & 24.00 & 29.25 & 41.25 & 48.25 & 44.50 \\
RW + Qwen-Image + REF & \textbf{73.75} & 32.25 & \textbf{46.75} & \textbf{49.25} & \textbf{47.75} \\ \bottomrule
\end{tabular}
}
\end{table}

\subsection{User Study}
To further validate the effectiveness of our proposed pipeline from a human perspective, we conducted a comprehensive user study. We randomly selected 20 table instances and generated visualizations using three configurations: the base generation model (Qwen-Image), the model prefixed with the rewriting module (RW+Qwen-Image), and our full ShowTable pipeline (RW + Qwen-Image + REF). We invited 20 human evaluators to independently assess the 20 sets of images. Participants were asked to vote for the best generation in each set across our five evaluation dimensions.
The preference rates are reported in \cref{tab:user_study}. Consistent with our quantitative automated metrics, the full ShowTable pipeline significantly outperforms the baselines in human preference. Notably, ShowTable achieves an overwhelming 73.75\% preference rate in Data Accuracy, confirming that our progressive reflection-refinement loop effectively corrects data-to-visual mapping errors that humans easily perceive. For Text Rendering (TR), the base model occasionally receives higher preference (38.50\%); this aligns with our observation, where the base model often defaults to merely rendering the raw table text directly without executing complex layout reasoning or data visualization. While this ``lazy rendering'' avoids text artifacts, it completely fails the core visualization task (reflected in its dismal 2.25\% DA score). Overall, the human study strongly corroborates that our pipeline produces the most factually faithful and visually appealing infographics.

\section{Limitations and Future Work}
While our ShowTable pipeline demonstrates significant improvements in creative table visualization, there remain several limitations that open avenues for future research:

\noindent\textbf{Full-pipeline training.} 
First, our current work primarily explores the training of the rewriting and refinement modules. The generation and reflection modules still rely on off-the-shelf models. We believe that extending supervised fine-tuning (SFT) or reinforcement learning (RL) to all components of the pipeline could further enhance the system's robustness and domain adaptability. Future work will investigate a more holistic training strategy to optimize the entire pipeline jointly.

\noindent\textbf{Towards a unified model.}
Second, the current implementation operates as a cascade of distinct models for rewriting, generation, reflection, and refinement. While effective, this multi-model approach increases deployment complexity and inference latency. A promising future direction is to explore a unified multi-modal architecture capable of performing all four sub-tasks end-to-end. Integrating these capabilities into a single model could significantly streamline the workflow and improve efficiency.

\noindent\textbf{Dependency on foundation models.}
Finally, the performance of our pipeline is inevitably constrained by the capabilities of the underlying base models. As observed in our refinement experiments, the upper bound of visualization fidelity is often determined by the precision of refinement module. We hope that our proposed task and benchmark will encourage the community to focus more on enhancing these foundational capabilities, particularly in precise fine-grained editing and logical reasoning, thereby advancing the field closer to more general and capable Artificial General Intelligence (AGI).


\end{document}